\documentclass[sigconf]{acmart}
\usepackage{amsfonts}
\usepackage{bm}
\usepackage{enumitem}
\usepackage[table]{xcolor}
\usepackage[utf8]{inputenc}
\usepackage{geometry}
\usepackage{pifont}
\usepackage{booktabs}  
\usepackage{multirow}  
\usepackage{graphicx}  
\usepackage{booktabs, tabularx, pifont, multirow}
\usepackage{tcolorbox}
\usepackage{amsmath}
\usepackage{xcolor}
\usepackage{caption}   
\usepackage{booktabs}  
\usepackage{array} 
\usepackage{caption}
\usepackage{makecell}
\usepackage{subcaption}
\newcolumntype{C}{>{\centering\arraybackslash}c}
\AtBeginDocument{%
  }

\graphicspath{{figures/}}

\usepackage{tcolorbox}

\newtcolorbox{promptbox}[1]{
  colback=gray!5!white,      
  colframe=black!70!white,   
  fonttitle=\bfseries,       
  coltitle=white,           
  boxrule=0.8pt,             
  arc=2mm,                   
  left=2mm,right=2mm,top=0mm,bottom=0mm,
  title={#1}                
}

\newtcolorbox{PromptDesign}[1]{
    enhanced,
    frame hidden,                   
    toprule=1.5pt,                  
    bottomrule=1.5pt,               
    colframe=black,                
    colback=white,                 
    sharp corners,                 
    boxsep=2pt,
    left=0pt, right=0pt, top=8pt, bottom=8pt, 
    title={#1},                     
    coltitle=black,                 
    fonttitle=\bfseries\large,     
    attach boxed title to top left={yshift=-12pt, xshift=0pt}, 
    boxed title style={frame hidden, colback=white}, 
    before skip=20pt,               
    after skip=20pt                 
}

\newcommand{\role}[1]{\textbf{\textsf{\color{blue!40!black}#1}}}

\setcopyright{acmlicensed}
\copyrightyear{2018}
\acmYear{2018}
\acmDOI{XXXXXXX.XXXXXXX}
\acmConference[Conference acronym 'XX]{Make sure to enter the correct
  conference title from your rights confirmation email}{June 03--05,
  2018}{Woodstock, NY}
\acmISBN{978-1-4503-XXXX-X/2018/06}

\acmSubmissionID{6450}

\renewcommand\footnotetextcopyrightpermission[1]{} 
\begin{document}

\title[Multimodal Conversational Stance Flipping Forecasting]{StanceFlip: A Comprehensive Multi-Dimensional Benchmark for Multimodal Conversational Stance Flipping Forecasting}


\author{Heyan Chai}
\email{hychai@szu.edu.cn}
\affiliation{%
  \institution{College of Computer Science and Software Engineering, Shenzhen University}
  \city{Shenzhen}
  \country{China}
}

\author{Xin Li}
\email{2024221014@mails.szu.edu.cn}
\affiliation{%
	\institution{College of Computer Science and Software Engineering, Shenzhen University}
	\city{Shenzhen}
	\country{China}
}
\author{Wenjie Wang}
\email{2024280356@mails.szu.edu.cn}
\affiliation{%
	\institution{College of Computer Science and Software Engineering, Shenzhen University}
	\city{Shenzhen}
	\country{China}
}
\author{Jianyang Qin}
\email{hychai@szu.edu.cn}
\affiliation{%
	\institution{Harbin Institute of Technology}
	\city{Shenzhen}
	\country{China}
}
\author{Chaoyang Li}
\email{hychai@szu.edu.cn}
\affiliation{%
	\institution{Harbin Institute of Technology}
	\city{Shenzhen}
	\country{China}
}
\author{Lu Wang}
\email{wanglu@szu.edu.cn}
\affiliation{%
	\institution{Shenzhen University}
	\city{Shenzhen}
	\country{China}
}
\author{Hao Chen}
\email{sundaychenhao@gmail.com}
\affiliation{%
	\institution{City University of Macau}
	\city{Macao SAR}
	\country{China}
}
\author{Qing Liao}
\authornote{Corresponding author: Qing Liao. Email: Liaoqing@hit.edu.cn..}
\email{liaoqing@hit.edu.cn}
\affiliation{%
	\institution{Harbin Institute of Technology}
	\city{Shenzhen}
	\country{China}
}

\newtcolorbox{acadbox}[1]{
  colback=gray!5!white,       
  colframe=black,             
  boxrule=0pt,                
  toprule=1pt,                
  bottomrule=1pt,             
  arc=0mm,                    
  left=4mm, right=4mm, top=4mm, bottom=4mm,
  title={},                   
  fontupper=\small,           
  before upper={\textbf{\textit{#1}}\par\vspace{3pt}} 
}

\renewcommand{\shortauthors}{Trovato et al.}

\begin{abstract}
Conversational stance detection has shifted from static text analysis to dynamic multimodal modeling. 
However, existing benchmarks exhibit three key limitations: failure to capture the dynamic evolution of beliefs, particularly during stance reversals; difficulty in disentangling affective states from logical reasoning; and neglect of the critical role of multimodal cues in resolving pragmatic ambiguities such as sarcasm. To address these limitations, we propose \textbf{StanceFlip}, a benchmark designed for multimodal conversational stance flipping forecasting over multi-turn dialogues across five modalities and multi-scenarios, which includes two novel subtasks: 1) \textbf{Multimodal Stance Sextuple Extraction}, extracting \textit{holder, target, emotion, sentiment, stance, and rationale} as static state snapshots of dialogue to capture fine-grained cognitive structures. 2) \textbf{Dynamic Stance Flip Attribution}, tracking stance reversals across the conversation and identifying their underlying triggers.
Alongside the dataset, we propose a dedicated framework, named \textbf{ConStaFF}, for Multimodal Conversational Stance Flipping Forecasting (MCSFF). Built upon a large language model, ConStaFF performs end-to-end stance reasoning, with a Thought-of-Stance (ToS) reasoning framework and a self-reflective verification mechanism integrated for structured stance modeling and faithful flip attribution. Specifically, ToS decomposes the reasoning process into specialized cognitive personas to formulate target propositions, resolve cross-modal conflicts, and infer historical stance trajectories, while self-reflective verification iteratively refines generated rationales against multimodal evidence to reduce causal hallucination. Extensive experiments show that our approach achieves state-of-the-art performance on both sextuple extraction and flip-trigger attribution, outperforming strong multimodal large language model baselines by substantial margins.
\end{abstract}



\received{20 February 2007}
\received[revised]{12 March 2009}
\received[accepted]{5 June 2009}

\maketitle

\section{Introduction}
Stance detection aims to identify user's attitude toward a specific target in discourse \cite{mohammad2016stance,sridhar2015joint}. Early studies primarily focused on binary stance classification from isolated utterances \cite{wei2018dynamic,liang2022zeroshot,ChaiCTDLFL25}. The task has since evolved to multi-turn conversational settings, where opinions are expressed, negotiated, and sometimes reversed through interaction \cite{li2023ctsdt,Ding2025ZSCsd}. More recently, the rise of multimedia platforms has pushed stance analysis beyond text-only conversations toward multimodal discourse, in which non-textual signals provide essential context for understanding speaker intent~\cite{niu2024mmmtcsd,wang2024multiclimate}. In such scenarios, multimodal signals, including images, audio, stickers, and video, provide essential context that text alone cannot convey, particularly when speakers use sarcasm, irony, or passive aggression to mask their true intent \cite{liang2024mmsd,niu2024mmmtcsd}. As a result, the field increasingly needs fine-grained structured representations that go beyond simple label assignment and instead support reasoning about why a speaker holds or changes a stance.
\begin{figure}[t!]  
	\centering    
	\includegraphics[width=\columnwidth]{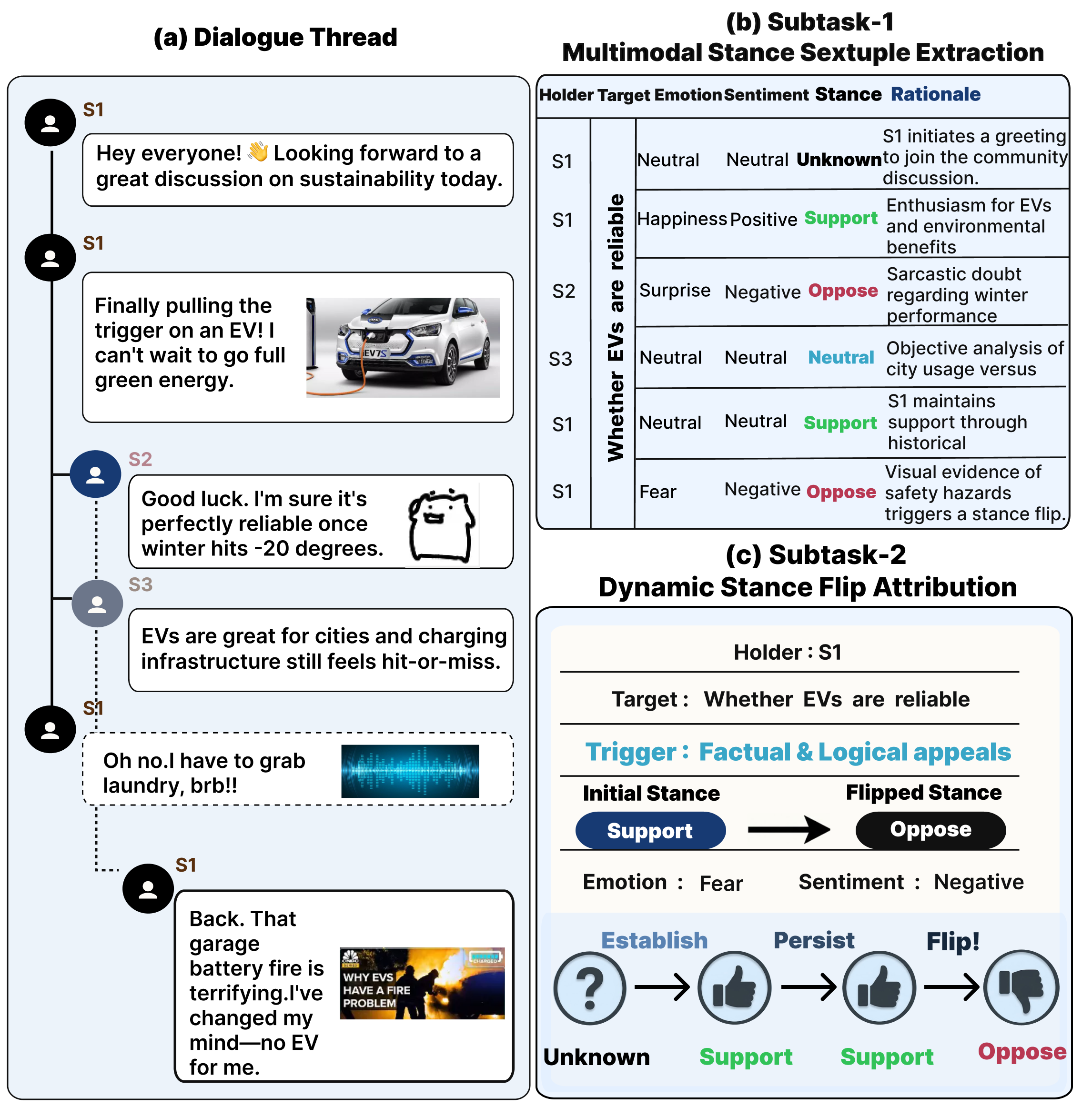} 
	\caption{Illustration of the StanceFlip benchmark.}
	\label{fig:intro}
\end{figure}

Despite this progress, existing research still leaves three important gaps. First, current methods assign a single polarity label to each utterance without disentangling the holder's affective expression from their target-specific stance. As the dialogue in Figure~\ref{fig:intro}(a) shows, Speaker \textit{S2} attaches a sarcastic sticker to express opposition despite the neutral surface text—since emotion and stance are different dimensions, one can happily oppose or angrily support. Without separate fields for emotion, sentiment, and stance anchored to the same holder-target pair, models inevitably conflate affective signals with logical stance. Second, most benchmarks treat non-textual information as auxiliary features rather than decisive evidence \cite{ChaiTCDF022,chen2024aspectsanchors}. In the same example, the video showing battery fire hazards is the factual trigger that causes Speaker \textit{S1} to reverse a previously firm stance, yet text-only or text-dominant pipelines cannot capture this cross-modal relationship \cite{wei2018dynamic,ChaiCTDLFL25}. Third, existing methods classify each turn independently and ignore the temporal continuity of stance \cite{niu2024mtcsd,Ding2025ZSCsd}. Stance typically exhibits persistence and tends to remain stable once established unless strong evidence leads to a change. To the best of our knowledge, existing benchmarks neither model the full process of stance evolution nor identify the factors that trigger stance changes.

To fill these gaps, we introduce StanceFlip, a large-scale bilingual benchmark for multimodal conversational stance flipping forecasting, comprising 7,710 annotated dialogues and 37,836 turns across five modalities and over 100  scenarios, with a flip density exceeding 20\%. Built from real-world conversational seeds, StanceFlip enriches multi-turn, multi-party dialogues with multimodal cues through GPT-4o-based synthesis and grounded retrieval, followed by dual-consistency filtering and expert verification. 
To disentangle affective expression from argumentative stance, we define two complementary subtasks. \textbf{Multimodal Stance Sextuple Extraction (subtask-I)} extracts a panoptic sextuple $(h, g, e, s, st, r)$ binding the \textit{holder, target, emotion, sentiment, stance, and rationale} into a single record, decoupling affective expression from argumentative stance in Figure~\ref{fig:intro}(b).\textbf{Dynamic Stance Flip Attribution (subtask-II)} tracks stance changes across conversations and attributes their underlying causes in Figure~\ref{fig:intro}(c). Together, the two subtasks advance stance detection from static label assignment to cognitively grounded causal reasoning.

Compared with conventional stance detection, the proposed task is more challenging, requiring models to track stance dynamics over multi-turn contexts, resolve cross-modal conflicts, and identify stance-change triggers. To address this, we propose \textbf{ConStaFF}, a dedicated framework built upon a multimodal large language model. To compensate for vanilla MLLMs' lack of structured mechanisms for target grounding, affective analysis, and temporal stance tracking, we design a novel \textbf{Thought-of-Stance (ToS)} reasoning framework that decomposes the task into four sequential cognitive steps handled by specialized expert personas, augmented with a self-reflective verification mechanism to reduce hallucinations and improve rationale faithfulness. Evaluations on StanceFlip show that ConStaFF consistently outperforms strong LLM baselines across both subtasks and languages.

In summary, the contributions of this work are threefold:
\begin{itemize}
	\item We formalize \textbf{Multimodal Conversational Stance Flipping Forecasting} via two subtasks---Multimodal Stance Sextuple Extraction and Dynamic Stance Flip Attribution---advancing stance detection from static label assignment to cognitively grounded causal reasoning.
	\item We contribute \textbf{StanceFlip}, a large-scale bilingual benchmark of 7,710 dialogues and 37,836 annotated turns across five modalities, featuring flip density exceeding 20\%, expert-level fidelity, and diverse domain coverage.
	\item We propose ConStaFF, an advanced framework equipped with the ToS reasoning framework and self-reflective verification mechanism, achieving state-of-the-art performance on both sextuple extraction and flip-trigger attribution, establishing a strong baseline for future work.
\end{itemize}

\section{Related Work}
\label{sec:related_work}

Stance detection was initially studied in isolated settings, where the goal is to infer a speaker’s attitude toward a target from a single utterance or post \cite{mohammad2016stance,aldayel2020survey,sridhar2015joint,zarrella2016mitre}. Subsequent work extended this setting to more challenging variants, including multiple targets \cite{sobhani2017dataset,wei2018dynamic}, zero-shot or unseen targets \cite{allaway2020zeroshot,liang2022zeroshot,allaway2021adversarial}, and target extraction in open-world settings \cite{li2023targetstance}. Despite these advances, the dominant formulation remains utterance-centric, treating stance as a local property of individual posts rather than a discourse-level state.

Recent work has therefore shifted toward conversational and multi-turn stance analysis, where opinions are expressed, negotiated, and revised through interaction \cite{li2023ctsdt,niu2024mtcsd,Ding2025ZSCsd}. This setting introduces substantially richer interaction patterns, including speaker alternation, topic drift, and target switching, making coarse-grained formulations increasingly inadequate. As a result, stance understanding requires finer-grained structure that explicitly anchors the holder, target, and supporting evidence across turns \cite{luo2024panosent,zheng2021mre,chen2024aspectsanchors}. However, existing formulations still often fail to separate a holder’s affective state from their argumentative stance, which leads to systematic ambiguity in multi-party dialogue and weakens stance attribution across turns \cite{niu2024mtcsd,Ding2025ZSCsd}.

In parallel, stance analysis has moved beyond text-only input toward multimodal settings \cite{liang2024mmsd,niu2024mmmtcsd,wang2024multiclimate}. This shift is crucial because non-textual signals can determine pragmatic meaning, override the polarity suggested by literal text, and reveal target-relevant evidence that remains implicit in language \cite{liang2024mmsd,niu2024mmmtcsd,wang2024multiclimate,hazarika2020misa,liang2021mmsarcasm}. Multimodality is therefore important not only for recognizing stance, but also for explaining stance reversals. Although prior work has incorporated discourse history and improved interpretability through logic or rationale generation \cite{zubiaga2017discourse,li2023ctsdt,lee2024logic,yuan2024reasoner,ma2024chain}, existing approaches still largely predict stance turn by turn rather than modeling it as a persistent state whose changes are triggered by new evidence. Related advances in adaptive conversational context modeling, modality-context disentanglement, consequence forecasting, emotion-cause generation, and causal or counterfactual debiasing further suggest that multimodal interaction understanding benefits from explicit modeling of latent causes, confounders, and future consequences \cite{chen2021drop,li2023disentanglecmer,ju2024ecfcon,wang2024observe,sun2022counterfactualmsa,sun2023gdebiasmsa}. Against this backdrop, we introduce \textit{StanceFlip}, a novel benchmark,  and \textit{ConStaFF},  reasoning framework that unifies fine-grained stance structure, multimodal grounding, and stance-flip modeling in a single task setting. Table~\ref{tab:comparison} summarizes the key differences between \textit{StanceFlip} and existing stance benchmarks.


\definecolor{MacaronBlue}{RGB}{140, 156, 177}
\definecolor{MacaronGreen}{RGB}{144, 165, 131}
\definecolor{MacaronYellow}{RGB}{209, 177, 141}
\definecolor{MacaronPink}{RGB}{167, 136, 129}
\definecolor{HeaderGray}{RGB}{240, 240, 240}

\begin{table*}[t]
\centering
\tiny
\renewcommand{\arraystretch}{0.6} 
\setlength{\tabcolsep}{3pt}       
\setlength{\abovecaptionskip}{0pt}   
\setlength{\belowcaptionskip}{0pt}   
\caption{Systematic comparison of StanceFlip with existing stance detection benchmarks.}
\label{tab:comparison}
\resizebox{\textwidth}{!}{
\begin{tabular}{lcccccccc}
\toprule
\rowcolor{HeaderGray} & \multicolumn{3}{c}{Context \& Modality} & \multicolumn{2}{c}{Granularity \& Scope} & \multicolumn{3}{c}{Cognitive Reasoning} \\
\cmidrule(lr){2-4} \cmidrule(lr){5-6} \cmidrule(lr){7-9}
\rowcolor{HeaderGray} Benchmark & Interaction & Modality & Language & Target Scope & Structural Elements & \makecell{Stance State \\ Machine} & \makecell{State \\ Rationale} & \makecell{Causal \\ Trigger} \\ 
\midrule
SemEval-16 \cite{mohammad2016stance} & Single-turn & Pure Text & EN & 5 Targets & (Target, Stance) & \ding{55} & \ding{55} & \ding{55} \\
VAST \cite{allaway2020zeroshot} & Single-turn & Pure Text & EN & Multiple Targets & (Target, Stance) & \ding{55} & \ding{55} & \ding{55} \\
\midrule
\rowcolor{MacaronBlue!15} CTSDT \cite{li2023ctsdt} & Multi-turn & Pure Text & EN & Single Target & (Target, Stance) & \ding{55} & \ding{55} & \ding{55} \\
\rowcolor{MacaronBlue!15} MT2-CSD \cite{niu2025mt2csd} & Multi-turn & Pure Text & EN & Multiple Targets & (Target, Stance) & \ding{55} & \ding{55} & \ding{55} \\
\midrule
\rowcolor{MacaronGreen!15} MmMtCSD \cite{niu2024mmmtcsd} & Multi-turn & Text + Image & EN, ZH & Multiple Targets & (Target, Stance) & \ding{55} & \ding{55} & \ding{55} \\
\rowcolor{MacaronGreen!15} MultiClimate \cite{wang2024multiclimate} & Multi-turn & Text + Video & EN & Single Target & (Target, Stance) & \ding{55} & \ding{55} & \ding{55} \\
\midrule
\rowcolor{MacaronPink!25} StanceFlip & Multi-turn & T+I+V+A+S* & EN, ZH & Multiple Targets & \makecell{Sextuple$^\dagger$ Stance Attribution$^\ddagger$} & \checkmark & \checkmark & \checkmark \\
\bottomrule
\end{tabular}
}
\begin{flushleft}
\scriptsize * Modality: T: Text, I: Image, V: Video, A: Audio, S: Stickers \& Memes. \\
\scriptsize $^\dagger$ Sextuple: (Holder, Target, Sentiment, Emotion, Stance, Rationale); $^\ddagger$ Stance-Flip Attribution: (Holder, Target, Stance, Flip, Trigger, Flip-Trig).
\end{flushleft}
\end{table*}

\begin{table}[t]
	\footnotesize
	\centering
	\caption{Detailed statistical profile of the StanceFlip benchmark.}
	\vspace{-0.2em}
	\label{tab:detailed_stats}
	\setlength{\tabcolsep}{0.8mm}{
		\begin{tabular}{lcccccccccccc}
			\toprule
			\multirow{2}{*}{Lang.} & \multicolumn{3}{c}{Scale and Depth} & & \multicolumn{2}{c}{Evo. Logic} & & \multicolumn{5}{c}{Multimodal Coverage (\%)} \\
			\cmidrule{2-4} \cmidrule{6-7} \cmidrule{9-13}
			& Dial. & Turn. & Av.  & & Sext. & Flip. & & Img. & Mem. & Aud. & Vid. & All \\
			\midrule
			EN & 3,898 & 21,534 & 5.52 & & 21,534 & 23.99\% & & 27.60 & 37.61 & 20.52 & 19.04 & 62.85 \\
			ZH& 3,812 & 16,302 & 4.28 & & 16,302 & 16.53\% & & 42.65 & 42.58 & 2.68 & 10.18 & 72.69 \\
			\midrule
			Tot/Avg & 7,710 & 37,836 & 4.91 & & 37,836 & 20.30\% & & 35.04 & 40.06 & 11.70 & 14.66 & 67.72 \\
			\bottomrule
		\end{tabular}
	}
\end{table}

\section{StanceFlip Benchmark}
In this section, we present StanceFlip, a large-scale multimodal benchmark designed to study the dynamic evolution of stance in conversational discourse. We describe the automated construction pipeline and summarize main  characteristics of  resulting dataset.

\subsection{Task Definition}
\label{subsec:task_definition}

We formally define the StanceFlip task as tracking the fine-grained evolution of holder stance as a dynamic, persistent state rather than as a series of isolated classification points. 
\begin{definition}[StanceFlip Task]
Let ${D} = \{T_1, T_2, \dots, T_n\}$ be a multimodal dialogue of $n$ turns. Each turn $T_i = (u_i, h_i, a_i)$ consists of a textual utterance $u_i$, a speaker $h_i$, and an associated multimodal information $a_i \in \{I^{img}, I^{meme}, I^{aud}, I^{vid}, \text{N/A}\}$. Given a specific debate proposition $G$, which serves as the consistent target $g$ for each turn, the system tracks the evolution of stances $st$ for each participant.
\end{definition}

\textbf{Stance Evolution Logic. }Unlike traditional stance detection that treats utterances independently, our benchmark assumes a holder's conviction is a continuous state. We establish the following transition principles for the stance label $st_i \in \{S, N, O, U\}$ (where $S, N, O, U$ are definitive states, namely \textit{Support}, \textit{Neutral}, \textit{Oppose}, and \textit{Unknown}):
\begin{itemize}[leftmargin=2.5em, labelsep=0.5em, nosep]
    \item \textbf{Stance Establishment:} A user forms a clear stance (e.g., \textit{Support} / \textit{Oppose}) after previously having no explicit opinion. Note that the transition from Unknown to a definitive state is called "Establishment," not a "Flip."
    \item \textbf{Stance Persistence:} Once a stance is formed, it remains unchanged across subsequent turns unless explicitly updated. Irrelevant or off-topic utterances inherit the previously expressed stance $st_{prev}$.
    \item \textbf{Stance Flip: }A stance flip occurs when a user changes from one clear stance to another (e.g., from \textit{Support} to \textit{Oppose}), which is the main focus of our task.
\end{itemize}

Drawing inspiration from panoptic sentiment analysis \cite{luo2024panosent}, we divide the task into two core subtasks.
\textbf{Subtask-I: Multimodal Stance Sextuple Extraction.} This subtask extracts a stance sextuple $S_i = (h, g, e, s, st, r)$ for each turn $T_i$. The elements include: (1) Holder $h$ and Target $g$; (2) Emotion $e$ (7 classes) and Sentiment $s$ (3 classes); (3) Stance $st$; and (4) Rationale $r$. These sextuples provide state-aware snapshots of the conversation.
\textbf{Subtask-II: Dynamic Stance Flip Attribution.} 
This subtask identifies turns where a stance flip occurs and attributes the flip to its causal trigger $r$. Specifically, given a dialogue ${D} = \{T_1, T_2, \dots, T_n\}$, 
the task conducts two steps: (1)\textit{Flipping Forecasting.} Identify turn $t$ where $st_t \neq st_{t-1}$ and $st_t, st_{t-1} \in \{S, N, O\}$. 
(2)\textit{Trigger Categorization.} Classify the trigger mechanism into one of four types: 1) Factual \& Logical Appeals, 2) Emotional \& Value-based Appeals, 3) Personal Experience \& Anecdote, or 4) Social Influence.
This task focuses on the \emph{temporal dynamics} and \emph{dynamic reasoning} of stance evolution, complementing Subtask-I's element extraction with attribution analysis.

\subsection{Automated Construction Pipeline}

To address the scarcity of multimodal dialogues with complex stance transitions, we propose a multi-stage simulation-retrieval pipeline that turns textual seeds into multimodal discourse. A detailed description is provided in \textbf{Appendix A} of supplementary materials.

\textbf{Data Sourcing and Standardization.} We curate textual seeds from three corpora: DailyDialog~\cite{Li2017DailyDialog}, MELD~\cite{Poria2019MELD}, and ZS-CSD~\cite{Ding2025ZSCsd}. Dialogues with 3--7 turns are selected to ensure sufficient context for stance evolution modeling, then mapped into a unified schema preserving speaker identities and conversation structure.

\textbf{Directed Multimodal Augmentation.} GPT-4o~\cite{openai2024gpt4o} identifies critical turns (e.g., shifts or core arguments) as injection points via stance significance scoring. An online chat decision tree then aligns the medium with the holder's intent: audio for paralinguistic cues, emojis/memes for emotional feedback, and images/videos for visual evidence.

\textbf{Multimodal Query Synthesis and Grounded Retrieval. } GPT-4o synthesizes queries emphasizing concrete sensory details, and generates cross-modal conflicts (e.g., pairing a 
compliment with an eye-rolling meme) to simulate irony and passive-aggression. Queries are encoded with SentenceTransformer~\cite{Reimers2019SBERT} to retrieve from COCO~\cite{Lin2014COCO}, our proprietary sticker dataset, AudioSet~\cite{Gemmeke2017AudioSet}, and WebVid~\cite{Bain2021WebVid}, selecting candidates with similarity $\geq 0.8$.
\begin{figure*}[t!] 
	\centering
	\includegraphics[width=\textwidth]{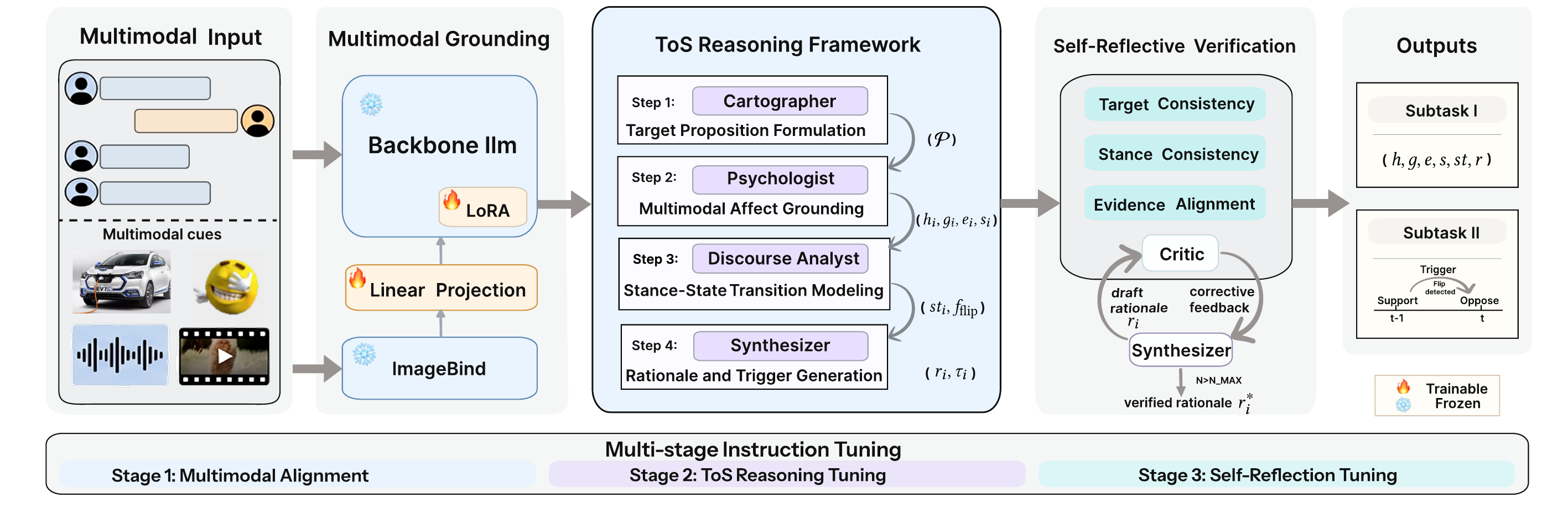} 
	\caption{The overall architecture of our ConStaFF model.}
	\label{fig:Model}
\end{figure*} 

\textbf{Evolution-aware Automated Labeling. } GPT-4o performs a panoptic scan based on our evolution logic, prioritizing multimodal evidence to decode latent conviction under irony. This yields per-turn logic-based rationales with keywords, emotions, and sentiment labels.

\subsection{Quality Assurance}
The reliability of StanceFlip rests on rigorous construction protocols and exhaustive manual verification.

\textbf{Construction Protocols and Dual-Consistency Filtering.} Structural heuristics enforce contextually justified injection and discourse-level consistency. A cross-modal grounding filter (similarity $\geq 0.8$) and a stance confidence gate (self-assessed score $\geq 0.9$) ensure rationale-label alignment.

\textbf{Expert-level Manual Review and Refinement.} To ensure data fidelity, the entire dataset underwent a comprehensive manual review and correction phase. Expert annotators inspected every dialogue thread to verify the coherence of stance trajectories, relevance of retrieved information, and accuracy of rationales. Errors in automated labeling or cross-modal misalignments were manually corrected, ensuring the benchmark meets expert-level standards.

\textbf{Case Study Validation.} A case study on 100 dialogues (${\sim}600$ turns) yielded Cohen's $\kappa = 0.84$ for stance identification and $0.87$ for multimodal relevance~\cite{landis1977measurement}. Against human gold standard,  automated pipeline achieved 86.4\% accuracy for stance detection and 88.5\% for multimodal matching.

\subsection{Data Highlights}
The StanceFlip benchmark is partitioned into train/validation/test splits at a ratio of 8:1:1, stratified by dialogue topic and language to ensure balanced target and flip-density distributions across splits. The statistics are shown in Table~\ref{tab:detailed_stats}. Below we summarize the key characteristics and highlights of the StanceFlip dataset:

\textbf{Panoptic Granularity and Cognitive Inference.} StanceFlip established a high standard for fine-grained analysis by extracting a complete sextuple $\mathcal{S} = (h, g, e, s, st, r)$ for every turn. We introduce natural language rationales that bridge multimodal cues with the holder's convictions, making the reasoning path easier to trace.

\textbf{Holistic Lifecycle Tracking of Stance States.} We formalize the stance lifecycle by tracking the trajectory from initiation to contextual inheritance and substantive inversion. This state-aware modeling captures the cumulative nature of discourse and turns stance detection into a dynamic sequence modeling task.

\textbf{Strategic Multimodal Synergy via Intent Alignment.}  Guided by an online chat decision tree, multimodal information is inserted based on communicative intents. By explicitly modeling polarity discrepancies (e.g., pairing compliments with ironic memes), we simulate social phenomena like irony and passive-aggression.

\textbf{High-Volatility Dynamics and Anti-Static Bias.} To address the  static bias in existing datasets, StanceFlip exhibits a flip density above 20\%. This forces models to capture subtle turning points and triggers rather than simply fitting majority-class static labels.

\textbf{Bilingual Equilibrium with Expert-level Fidelity.} StanceFlip maintains a near 1:1 balance between English and Chinese. Combined with strict filtering and full manual verification, the benchmark achieves expert-level fidelity at scale.

\section{Methodology}
In this section, we present \textbf{ConStaFF}, a comprehensive framework tailored for multimodal stance flipping analysis shown in Figure~\ref{fig:Model}. Our framework is specifically designed to tackle the key challenges of this task, including complex conversational context understanding, multimodal information fusion, and cognitive-level stance flipping reasoning. We then introduce the overall architecture, the ToS reasoning framework, the self-reflective verification mechanism, and the multi-stage instruction tuning strategy.

\subsection{ToS Reasoning Framework}
Standard Chain-of-Thought (CoT) reasoning~\cite{CoT} treats inference as a single linear chain, while Tree of Thoughts (ToT)~\cite{ToT} emphasizes generic branching exploration. Neither directly addresses the core challenges of multimodal conversational stance-flip forecasting, namely consistent target grounding in multi-party dialogue, conflict resolution between textual and non-textual cues, discourse-level stance-state tracking, and evidence-grounded explanation. To address these challenges, we propose the \textbf{Thought-of-Stance (ToS)} framework, a stance-centered reasoning process tailored to our task. As shown in Figure~\ref{fig:Model}, rather than expanding arbitrary thought branches, ToS decomposes reasoning into four ordered roles: a \textit{Cartographer} for target proposition formulation, a \textit{Psychologist} for multimodal affect grounding, a \textit{Discourse Analyst} for stance-state transition modeling, and a \textit{Synthesizer--Critic} pair for explanation construction and verification. The hierarchy in ToS is therefore explicit: reasoning proceeds from global dialogue anchoring, to turn-level multimodal interpretation, to cross-turn stance updating, and finally to evidence-grounded explanation.

\textbf{Step 1: The Cartographer (Target Proposition Formulation).} 
To resolve target ambiguity caused by topic drift and speaker alternation, the model first acts as a ``Cartographer'' to perform a global scan of the dialogue and multimodal cues to formulate a debate proposition $\mathcal{P}$. This proposition serves as the semantic anchor for all subsequent steps, ensuring that emotion, sentiment, stance, and flip triggers are inferred with respect to the same target rather than different local mentions. This step is formulated as:
\begin{equation}
	\mathcal{P} \leftarrow f_{\text{cart}}(D, \mathcal{M} \mid \mathcal{I}_{1})
\end{equation}
where $D$ represents the dialogue history, $\mathcal{M}$ denotes multimodal features, and $\mathcal{I}_1$ is the instruction.
\begin{promptbox}{Step 1: Target Proposition Identification}
	\textbf{Input Data:} Full dialogue text with captions. \\
	\textbf{Instruction:} Act as a \role{[Cartographer]}. Identify the central `Target'. Scan the entire dialogue to identify the main topic of contention. Convert this core issue into a clear, concise debate proposition sentence (e.g., ``The debate concerning whether coffee should be tried as a substitute for cigarettes''). This proposition will serve as the consistent reference point. \\
	\textbf{Output:} (Target Proposition: [Debate Proposition Sentence])
\end{promptbox}

\textbf{Step 2: The Psychologist (Multimodal Conflict Resolution).} 
To resolve pragmatic ambiguity, the model acts as a ``Psychologist'' to ground the holder's affective state in multimodal evidence, and then to  identify the discrepancy  when textual polarity conflicts with non-textual cues or target-relevant media evidence, such as supportive wording paired with sarcastic prosody or a mocking sticker. Rather than treating multimodal signals as auxiliary features, this step uses them to determine the holder's emotion and sentiment toward $\mathcal{P}$.

\begin{promptbox}{Step 2: Multimodal Affect Grounding}
\textbf{Input Data:} Dialogue history, current speaker, utterance with cues, core target. \\
\textbf{Instruction:} Act as a \role{[Psychologist]} and \role{[Emotion Analyst]}. Infer the holder's emotion and sentiment toward the target by jointly considering text and multimodal cues. If the literal text conflicts with multimodal evidence, prefer the interpretation that best resolves the pragmatic ambiguity.  \\
\textbf{Output:} (Holder, Target, Emotion, Sentiment)
\end{promptbox}
 
This step can be formulated as:
\begin{equation}
    (h_i, g_i, e_i, s_i) \leftarrow f_{\text{psy}}(D, u_i, \mathcal{M}_i, \mathcal{P} \mid \mathcal{I}_{2})
\end{equation}
where $u_i$ is the current utterance, $\mathcal{M}_i$ is the multimodal evidence of the current turn, and $e_i$ and $s_i$ denote emotion and sentiment.

\begin{promptbox}{Step 3: Stance-State Transition Modeling}
	\textbf{Input Data:} Dialogue history, current speaker, current emotion/sentiment, previous stance, and target proposition. \\
	\textbf{Instruction:} Act as a \role{[Discourse Analyst]}. Determine the holder's current stance toward the target by jointly considering dialogue history, current affective evidence, and the previous stance. Mark a flip only when the evidence supports a substantive change in target-specific position. \\
	\textbf{Output:} (Holder, Target, Emotion, Sentiment, Stance, Flip)
\end{promptbox}
\textbf{Step 3: The Discourse Analyst (Stance Evolution).} 
The model acts as a ``Discourse Analyst'' to determine whether the current turn reflects stance persistence or stance change. In ToS, stance is treated as a discourse state rather than a turn-local label. Emotion and sentiment are used as intermediate evidence, but the final stance is inferred by comparing the current turn with the holder's prior stance state in context. A flip is detected only when the evidence supports a substantive revision of the holder's position toward $\mathcal{P}$.
We formulate this step as:
\begin{equation}
	(h_i, g_i, e_i, s_i, st_i, f_{\text{flip}}) \leftarrow f_{\text{ana}}(D, h_i, e_i, s_i, st_{prev}, \mathcal{P} \mid \mathcal{I}_{3})
\end{equation}
where $st_i$ is the current stance, $st_{prev}$ is the previous stance, and $f_{\text{flip}}$ is the flip indicator.

\begin{promptbox}{Step 4: Rationale and Trigger Generation}
	\textbf{Input Data:} Dialogue history, multimodal evidence, and intermediate results. \\
	\textbf{Instruction:} Act as a \role{[Synthesizer]}. Generate a concise explanation of why holder maintains or changes the stance, grounded in the most relevant textual and multimodal evidence. Identify the trigger that caused the transition when $f_{flip}=1$.\\
	\textbf{Output:} (Rationale, Trigger Type)
\end{promptbox}
\textbf{Step 4: The Synthesizer (Rationale and Trigger Generation).}
Finally, once the stance state is determined, the model acts a ``Synthesizer'' to generate a rationale explaining why  holder maintains or changes stance, and identifies the trigger $\tau_i$ when a flip is detected ($f_{flip}=1$); otherwise ($f_{flip}=0$), only the rationale is generated. This step summarizes the textual and multimodal evidence most relevant to the inferred stance transition, producing an initial explanation for subsequent verification. We formulate this as:
\begin{equation}
	( r_i, \tau_i) \leftarrow f_{\text{sys}}(D, h_i, g_i, s_i, e_i,st_{prev},st_i, f_{\text{flip}}, \mathcal{M}_i\mid \mathcal{I}_{4})
\end{equation}
where $r_i$ denotes the rationale and $\tau_i$ denotes the trigger type.

\begin{table*}[t!]
	\centering
	\caption{Main comparative results on subtask-I, Multimodal Stance Sextuple Extraction. "T/E/Se/St/R" represents Target, Emotion, Sentiment, Stance, and Rationale, respectively. All the scores are averaged over five runs under different random seeds.}
	\vspace{-10pt}
	\resizebox{\linewidth}{!}{%
		\begin{tabular}{llcccCCCCCCCCCC}
			\toprule
			\multirow{2}{*}{}  & \multirow{2}{*}{Model} & \multirow{2}{*}{Param.} & \multicolumn{6}{c}{Items} &   \multicolumn{4}{c}{Pairs}   &Sext. & Quad. \\
			\cmidrule[0.8pt](lr){4-9} \cmidrule[0.8pt](lr){10-13} \cmidrule[0.8pt](lr){14-14}\cmidrule[0.8pt](lr){15-15}
			&  &  &H& T & E & Se & St & R& T-E & T-Se & T-St & St-R& Micro. & Iden.  \\
			\midrule
			\multirow{10}{*}{English}  
			& Vicuna & 7B & 96.51 &45.55 & 45.22 & 38.84 & 48.91 & 32.12 & 26.75 & 22.05 & 32.34 & 22.16 & 6.04 & 18.69 \\
			& Llama2 & 7B & 83.86&28.28 & 42.03 & 41.65 & 27.63 & 12.47 & 13.88 & 13.50 & 9.38 & 10.28 & 0.64 & 3.86 \\
			& Llama3 & 8B &93.66 &43.14 & 59.64 & 57.42 & 44.66 & 25.96 & 29.53 & 26.26 & 22.70 & 20.62 & 4.90 & 11.28 \\
			& Qwen2.5 & 7B & 95.56&50.93 & 57.91 & 73.94 & 56.12 & 44.99 & 30.44 & 37.56 & 28.95 & 29.84 & 6.24 & 22.72 \\
			& Llama2 (ToS+self-reflection) & 7B & 99.93& 40.92 & 61.38 & 65.97 & 51.74 & 49.96 & 26.09 & 26.24 & 24.17 & 37.95 & 8.30 & 20.46 \\
			& Llama3 (ToS+self-reflection) & 8B &99.37 &47.81 & 63.55 & 61.47 & 67.56 & 60.43 & 29.99 & 26.97 & 33.85 & 48.70 & 11.43 & 28.36 \\
			& Flan-T5-XXL (ToS+self-reflection) & 11B & 99.69&45.43 & 63.99 & 74.68 & 59.69 & 63.55 & 31.03 & 37.56 & 31.63 & 42.46 & 14.85 & 29.84 \\
			& Qwen2.5 (ToS+self-reflection) & 7B & 96.74&64.03 & 66.67 & 66.52 & 62.51 & 44.54 & 44.69 & 42.61 & 42.76 & 37.71 & 12.18 & 28.80 \\
			& Mistral (ToS+self-reflection) & 7B & 99.28&62.66 & 68.80 & 73.94 & 56.12 & 51.37 & 45.14 & 47.07 & 36.82 & 38.60 & 13.36 & 30.44 \\
			& \textbf{ConStaFF (Ours)} & 7B &\textbf{99.93} &\textbf{64.88} & \textbf{73.05} & \textbf{74.68} & \textbf{75.43} & \textbf{72.01} & \textbf{48.55}  &\textbf{50.48} & \textbf{52.41} & \textbf{61.62} & \textbf{25.84} & \textbf{46.77} \\
			\midrule
			\multirow{8}{*}{Chinese} & qwen2.5 & 7B & 100&48.76 & 47.76 & 49.70 & 39.89 & 53.56 & 24.06 & 24.20 & 20.69 & 30.72 & 4.37 & 29.29 \\
			& Llama2 & 7B& 100 &23.01 & 49.51 & 51.48  &47.53 & 29.99 & 13.18 & 14.37 & 11.87 & 20.83 & 1.32 & 8.50 \\
			& Llama3 & 8B & 100 &26.98 & 48.44 & 62.61 & 38.21 & 55.38 & 13.74 & 17.89 & 10.73 & 27.98 & 2.08 & 16.46 \\
			& Vicuna (ToS+self-reflection) & 7B & 99.15&59.46 & 63.55 & 71.96 & 61.54 & 51.78 & 37.92 & 44.24 & 38.78 & 37.20 & 12.42 & 28.87 \\
			& Llama2 (ToS+self-reflection) & 7B & 99.68&48.11 & 68.15 & 73.03 & 47.97 & 51.20 & 32.46 & 36.34 & 23.70 & 36.27 & 7.97 & 25.35 \\
			& Llama3 (ToS+self-reflection) & 8B& 99.68 &53.14 & 54.99 & 71.02 & 49.26 & 56.88 & 33.97 & 36.77 & 25.92 & 39.35 & 10.41 & 31.45 \\
			& Mistral (ToS+self-reflection) & 7B & 99.67& 50.84 & 74.83 & 76.48 & 60.97 & 57.38 & 37.85 & 38.35 & 32.60 & 42.01 & 14.15 & 32.32 \\
			& \textbf{ConStaFF(Ours)} & 7B & \textbf{100}& \textbf{64.85} & \textbf{74.90} & \textbf{78.20} & \textbf{61.54} & \textbf{69.08} & \textbf{49.26} & \textbf{50.84} & \textbf{41.01} & \textbf{51.85} & \textbf{21.69} & \textbf{44.31} \\
			\bottomrule
		\end{tabular}%
	}\label{tab:task1_results}
\end{table*}
\subsection{Self-Reflective Verification Mechanism}
While ToS provides a structured process for target grounding, affective modeling, stance-state tracking, and initial rationale generation, the resulting explanation for stance transitions are particularly vulnerable to hallucination, as the model may generate plausible but unsupported reasons that are not fully grounded in the dialogue context or multimodal evidence. Motivation by~\cite{SelfRefine}, we introduce a novel self-reflective verification mechanism to improve explanation faithfulness, which verifies and refines the rationale generated by ToS rather than performing a new round of stance inference.

Specifically, we adopt a \textit{Synthesizer--Critic} loop. The \textit{Synthesizer} first produces a draft rationale based on the predicted stance state and the available dialogue context, while the \textit{Critic} checks whether this rationale is consistent with the target proposition, aligned with the inferred stance, and explicitly supported by the textual and multimodal evidence. If the rationale is vague, target-inconsistent, or unsupported by the evidence, the \textit{Critic} feeds corrective feedback back to the \textit{Synthesizer} for revision. Empirically, we find that the first refinement iteration yields the most significant improvement in rationale quality, while additional iterations provide diminishing returns at the cost of increased inference time. Therefore, we set $N_{\text{max}} = 1$ in this work, and the loop terminates either when all three verification criteria in $\mathcal{C}_i$ are satisfied or upon reaching $N_{\text{max}}$.

Formally, the verification process is defined as:
\begin{equation}
	r_i^{*} = \Phi_{\text{refine}}\Big(
	\underbrace{\Phi_{\text{syn}}(D_i, \mathcal{M}_i, st_i)}_{\text{Draft Rationale}},
	\mathcal{C}_i \mid \mathcal{I}_{\text{critic}}
	\Big)
\end{equation}
where $\Phi_{\text{syn}}$ generates an initial rationale draft for the inferred stance transition, and $\Phi_{\text{refine}}$ iteratively critiques and revises this draft for at most $N_{\text{max}}$ iterations. The constraint set $\mathcal{C}_i$ includes three criteria: 1)\textbf{Target Consistency}, requiring the rationale to remain grounded in the debate proposition; 2) \textbf{Stance Consistency}, requiring the explanation to support the predicted stance or stance reversal; and 3) \textbf{Evidence Alignment}, requiring explicit grounding in the dialogue content and, when applicable, the multimodal cues. The final output $r_i^{*}$ is a verified rationale that is more specific, evidence-grounded, and faithful to the stance prediction.

\begin{table}[t!]
	\centering
	\caption{Main comparative results on subtask-II.}
	\vspace{-10pt}
	\label{tab:task2_compact}
	\resizebox{1.0\linewidth}{!}{
		\begin{tabular}{llcccc} 
			\toprule
			Dataset & Model & Param.  & Flip & Trig & Flip-Trig \\
			\midrule
			\multirow{10}{*}{English} 
			& Vicuna & 7B & 17.49 & 28.14 & 9.89 \\
			& Llama2 & 7B & 19.23 & 5.77 & 5.77 \\
			& Llama3 & 8B & 11.11 & 15.87 & 7.94 \\
			& Qwen2.5 & 7B & 16.47 & 14.90 & 4.71 \\
			& Llama2 (ToS+self-reflection) & 7B & 21.37 & 32.82 & 15.27 \\
			& Llama3 (ToS+self-reflection) & 8B & 35.10 & 33.47 & 24.49 \\
			& Flan-T5-XXL (ToS+self-reflection) & 11B & 35.59 & 32.20 & 25.42 \\
			& Qwen2.5 (ToS+self-reflection) & 7B & 39.66 & 25.86 & 25.86 \\
			& Mistral (ToS+self-reflection) & 7B & 40.00 & 25.45 & 25.45 \\
			& \textbf{ConStaFF(Ours)} & 7B & \textbf{44.26} & \textbf{34.04} & \textbf{26.38} \\
			\midrule
			\multirow{8}{*}{Chinese} & Qwen2.5 & 7B & 13.95 & 16.28 & 11.63 \\
			& Llama2 & 7B & 17.46 & 17.46 & 9.52 \\
			& Llama3 & 8B & 12.20 & 9.76 & 9.76 \\
			& Vicuna (ToS+self-reflection) & 7B & 40.00 & 30.34 & 23.45 \\
			& Llama2(ToS+self-reflection) & 7B & 22.86 & 24.76 & 19.05 \\
			& Llama3(ToS+self-reflection) & 8B & 22.22 & 35.59 & 16.30 \\
			& Mistral(ToS+self-reflection) & 7B & 35.94 & 17.19 & 15.62 \\
			& \textbf{ConStaFF(Ours)} & 7B & \textbf{44.30} & \textbf{33.56} & \textbf{26.85} \\
			\bottomrule
		\end{tabular}%
	}
\end{table}
\subsection{Multi-stage Instruction Tuning}
A single-stage end-to-end objective tends to entangle multimodal grounding, structured stance reasoning, and explanation verification. To better instantiate the ToS framework and self-reflective verification, we adopt a progressive three-stage tuning strategy that equips ConStaFF with multimodal perception, structured stance reasoning, and self-correction ability.

\textbf{Stage 1: Multimodal Alignment (Perception).} 
We freeze the pre-trained ImageBind encoder~\cite{ImageBind} and train a lightweight linear projection layer on multimodal cue-description pairs from StanceFlip's training split, optimizing the language modeling loss on caption generation. This maps multimodal representations into the LLM embedding space, providing grounded inputs for subsequent ToS reasoning.


\textbf{Stage 2: ToS Reasoning Tuning (Cognition).} 
We fine-tune the model to follow the ordered ToS reasoning process (\textit{Proposition $\rightarrow$ Sentiment\&Emotion $\rightarrow$ Stance $\rightarrow$ Rationale}) using LoRA~\cite{LoRA} while keeping ImageBind fixed:
\begin{equation}
	W = W_0 + \Delta W, \qquad \Delta W = BA,
\end{equation}
where $W_0$ is the frozen pre-trained weight, $B \in \mathbb{R}^{d \times r}$ and $A \in \mathbb{R}^{r \times k}$ are trainable low-rank matrices, and $r \ll \min(d,k)$. This stage instills structured reasoning over target grounding, multimodal evidence interpretation, stance tracking, and rationale generation.


\textbf{Stage 3: Self-Reflection Tuning (Metacognition).} 
To mitigate hallucinations in trigger identification, we introduce a dedicated self-reflection tuning strategy by constructing supervision tuples consisting of a draft rationale, critique feedback, and a revised rationale. These tuples are synthesized by perturbing rationale drafts and training the model to first identify unsupported or inconsistent claims and then rewrite them into evidence-grounded explanations. Rather than introducing a new inference task, this stage directly trains the verification behavior described above and improves the faithfulness of rationale generation, especially for trigger identification. This stage improves the model's metacognitive ability to detect and revise reasoning errors, particularly benefiting  the ``Flip-Trigger'' metric.

\section{Experiments}
This section presents experiments to demonstrate the effectiveness of \textbf{ConStaFF} for stance flipping forecasting. We aim to answer the following research questions:
\begin{itemize}
    \item \textbf{RQ1} (Comparative Experiment) Can proposed ConStaFF improve the stance flipping forecasting performance ?  
    \item \textbf{RQ2} (Implicit Stance Experiment) How does {ConStaFF} perform in the implicit stance scenario?
    \item \textbf{RQ3} (Zero-shot Experiment) How does {ConStaFF} perform in the zero-shot stance flipping prediction scenario?
    \item \textbf{RQ4} How does the proposed ToS reasoning framework contribute to the performance?
    \item \textbf{RQ5} How significant is the role of multimodal information?
    \item \textbf{RQ6} Does the self-reflective verification mechanism contribute to our model?
\end{itemize}

\begin{figure}[t]
	\setlength{\abovecaptionskip}{0cm}
	\setlength{\belowcaptionskip}{0cm}
	\centering
	\subfloat[Stance]{
		\includegraphics[width=0.48\linewidth]{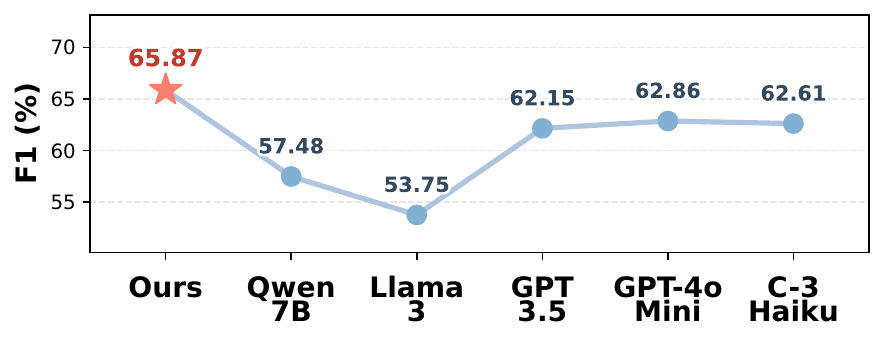}
	}
	\subfloat[Flip]{
		\includegraphics[width=0.48\linewidth]{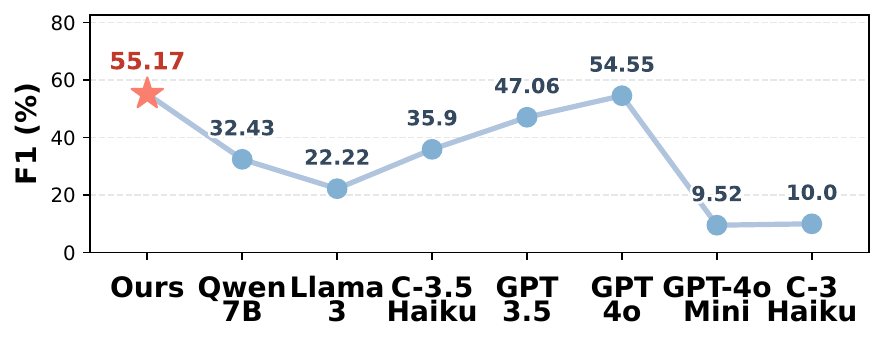}
	}\\
	\subfloat[Trigger]{
		\includegraphics[width=0.48\linewidth]{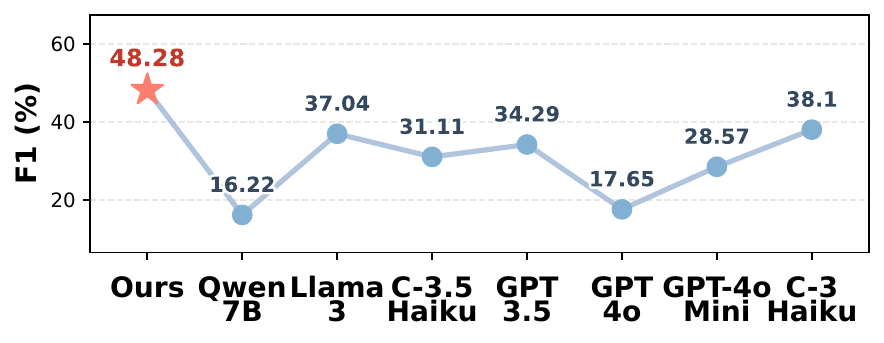}
	}
	\subfloat[Flip-Trigger]{
		\includegraphics[width=0.48\linewidth]{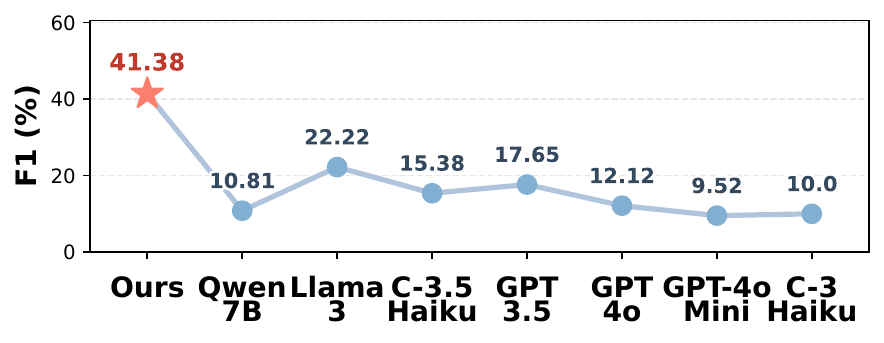}
	}
	\caption{Performance evaluation on implicit stance datasets. }
	\label{fig:res_implicit}
\end{figure}
\subsection{Experimental Settings}

\textbf{Evaluations.} We evaluate ConStaFF on StanceFlip across both subtasks. For Subtask-I, performance is assessed at three levels: item extraction, pair extraction, and full sextuple prediction. Target, Emotion, Sentiment, and Stance use F1-score; Rationale uses soft matching for semantic equivalence. A pair is correct only when both 
elements match, and a sextuple only when all components match; we report Micro-F1 on the complete sextuple and Identification F1 on the quadruple (Holder-Target-Stance-Flip). For Subtask-II, three F1-level metrics are reported: initial/flipped stance (Flip), trigger category (Trig), and flipped stance with trigger (Flip-Trig). All results are averaged over five runs with different random seeds.

\textbf{Baselines and Implementation.} 
We compare ConStaFF against strong MLLM baselines of different scales, including Flan-T5-XXL (11B)~\cite{FlanT5}, Vicuna (7B), Llama-2 (7B)~\cite{Llama2}, Llama-3 (8B)~\cite{Llama3}, Mistral (7B)~\cite{jiang2023mistral7b}, and Qwen2.5 (7B)~\cite{qwen25}. 
For fair comparison, ConStaFF adopts Vicuna-7B for English and Qwen2.5-7B~\cite{qwen25} for Chinese, both fine-tuned with LoRA under the ToS and self-reflection setting. The experiments were conducted on high-performance device with 8*Nvidia RTX A6000 GPUs. To ensure the reliability and reproducibility of our experiment, all results are averaged over five runs with different random seeds. Due to the space limitation, we provide the detailed description of model details and experimental setting in \textbf{Appendix C} of supplementary materials.

\subsection{Main Results (RQ1)}


\textbf{Performance on Multimodal Stance Sextuple Extraction task.}
Table~\ref{tab:task1_results} shows that ConStaFF delivers the strongest performance on Multimodal Stance Sextuple Extraction in both English and Chinese. The largest gains appear at the structure level: on English, ConStaFF improves Micro-F1 and Iden. to 25.84\% and 46.77\%, clearly above the strongest baselines (14.85\% and 30.44\%); on Chinese, the corresponding scores reach 21.69\% and 44.31\%, again establishing a clear margin over prior systems. The improvements are also consistent on difficult aspects such as Stance, Rationale, and the St-R pair, suggesting that our method enhances not only element recognition but also cross-element coherence within the full sextuple. More broadly, ToS+self-reflection variants consistently outperform their vanilla backbones, highlighting the effectiveness of our ConStaFF on structured reasoning for fine-grained stance modeling .

\noindent\textbf{Performance on Dynamic Stance Flip Attribution task.}
Table~\ref{tab:task2_compact} reports the results on Dynamic Stance Flip Attribution. ConStaFF achieves the best Flip and Flip-Trig scores in both languages, reaching 44.26\%/26.38\% on English and 44.30\%/26.85\% on Chinese. It also obtains the best Trig score on English and remains competitive on Chinese, where the strongest trigger-only baseline still falls short on the joint Flip-Trig metric. This indicates that ConStaFF is more effective at jointly modeling stance reversal and its underlying trigger, rather than optimizing the two objectives independently. At the same time, the relatively low joint scores across all systems confirm that flip attribution remains substantially more challenging than static stance prediction.
\begin{figure}[t!]
	\setlength{\abovecaptionskip}{0cm}
	\setlength{\belowcaptionskip}{0cm}
	\centering
	\subfloat[Bitcoin]{
		\includegraphics[width=0.48\columnwidth]{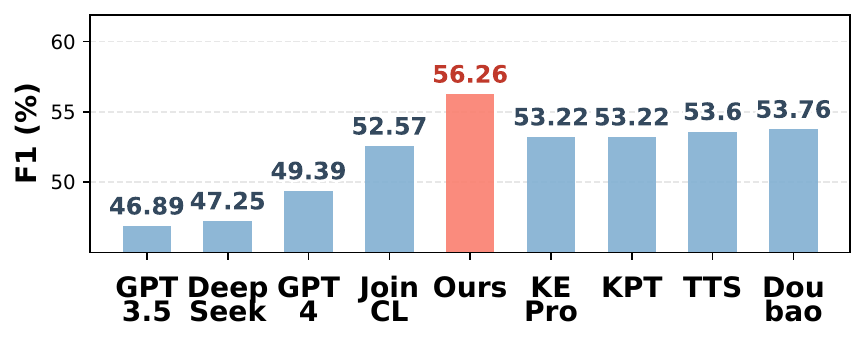}
	}
	\subfloat[SpaceX]{
		\includegraphics[width=0.48\columnwidth]{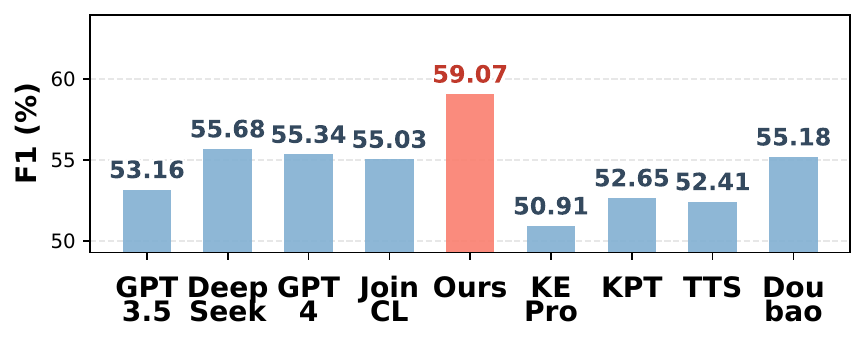}
	}	\\
	\subfloat[Bar\_Trump]{
		\includegraphics[width=0.48\columnwidth]{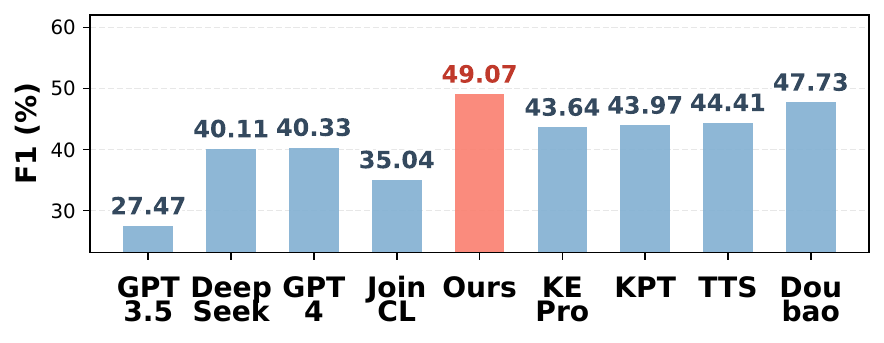}
	}
	\subfloat[Average]{
		\includegraphics[width=0.48\columnwidth]{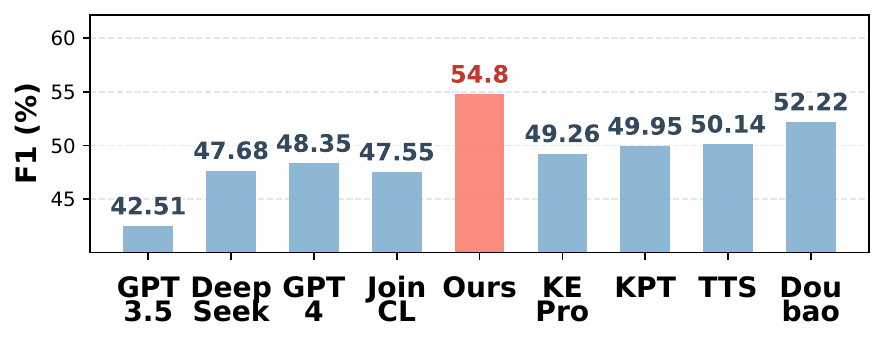}
	}
	\caption{Evaluation of the generalization to unseen targets.}
	\label{fig:res_unseen_target}
\end{figure}
\begin{figure}[t]
	\setlength{\abovecaptionskip}{0cm}
	\setlength{\belowcaptionskip}{0cm}
	\centering
	\includegraphics[width=0.99\linewidth]{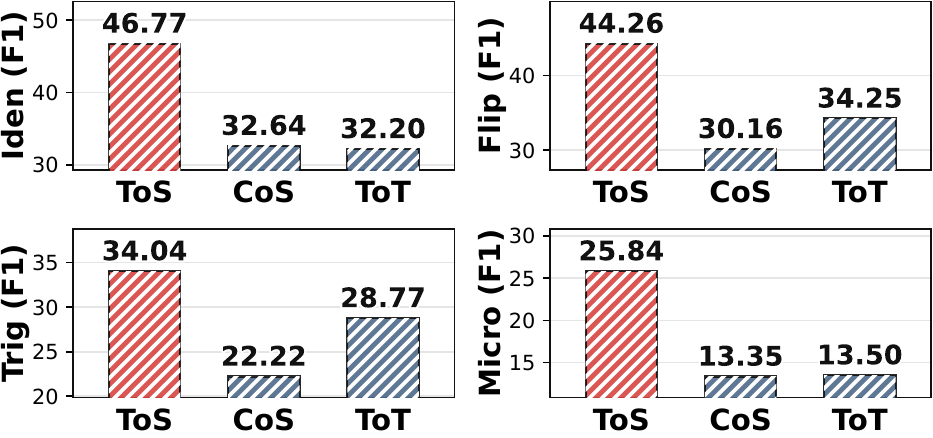}
	\caption{Evaluation of effectiveness of our ToS reasoning strategy.}
	\label{fig:abla_tos}
\end{figure}
\subsection{Performance on Implicit Stance (RQ2)}
Figure~\ref{fig:res_implicit} reports results on implicit stance cases involving sarcasm and visual metaphor, which require both correct stance recognition and evidence-grounded reasoning about why a stance flip occurs. A clear pattern is that strong LLM baselines remain competitive on the easier prediction targets, e.g., GPT-4o achieves 54.55\% on Flip, but their performance drops sharply on Trigger and especially on Flip-Trigger, suggesting that they often infer the outcome without accurately identifying its underlying cause. In contrast, ConStaFF achieves the best performance on all four metrics. The gains are especially pronounced on the reasoning-intensive metrics, where ConStaFF surpasses the strongest baseline by 10.18\% on Trigger and 19.16\% on Flip-Trigger. These results indicate that ConStaFF is more robust to implicit pragmatic cues and better aligns stance prediction with grounded trigger identification.

\subsection{Zero-Shot Generalization to Unseen Targets}
%

To evaluate zero-shot generalization, we test ConStaFF on the MT-CSD dataset~\cite{niu2024mtcsd} across three unseen target domains (\textit{Bitcoin}, \textit{SpaceX}, and \textit{Trump}), where ConStaFF is trained solely on StanceFlip without any exposure to MT-CSD during training. As shown in Figure~\ref{fig:res_unseen_target}, ConStaFF consistently ranks first across all three domains, achieving 56.26\% on \textit{Bitcoin}, 59.07\% on \textit{SpaceX}, and 49.07\% on \textit{Trump}, with the best average F1 of \textbf{54.80\%}. It outperforms the strongest baseline Doubao-Pro~\cite{doubao2024} by 2.58\% and the best transfer-learning method TTS~\cite{li2023tts} by 4.66\%. We attribute this to the ToS reasoning framework, which captures target-invariant stance evolution patterns rather 
than topic-specific lexical cues, enabling robust generalization to unseen targets.
\begin{figure}[t!]
	\centering
	\subfloat[Iden (F1)]{
		\includegraphics[width=.32\linewidth]{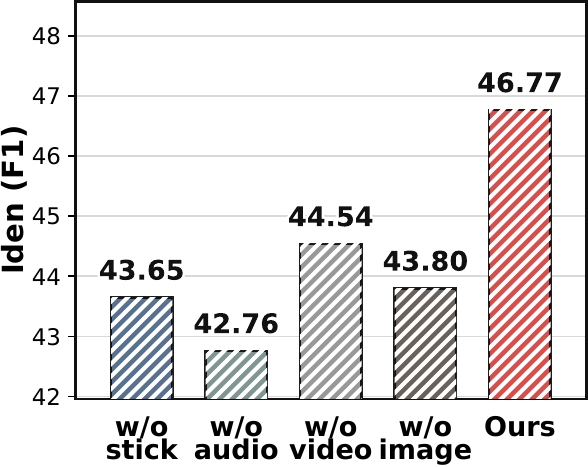}
	}
	\subfloat[Micro (F1)]{
		\includegraphics[width=.32\linewidth]{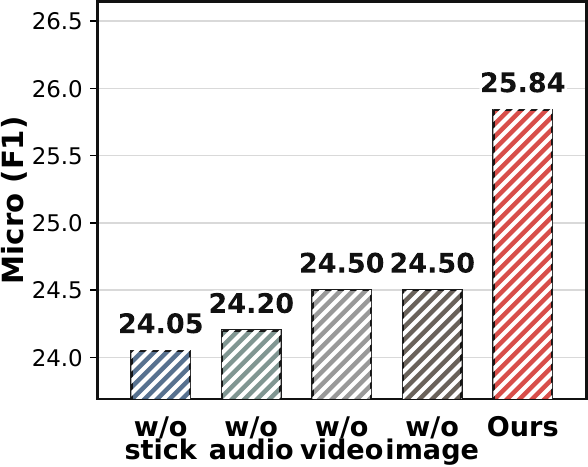}
	}
	\subfloat[Flip-Trig (F1)]{
		\includegraphics[width=.32\linewidth]{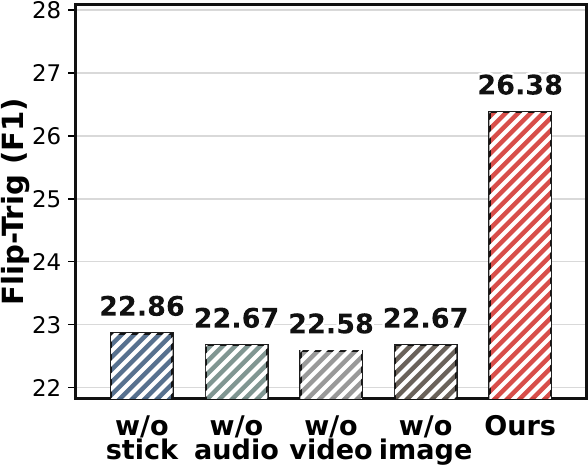}
	}
	\caption{Evaluation of the contribution of each modality.}
	\label{fig:abla_multimodal}
\end{figure}
\subsection{Effectiveness of ToS Reasoning (RQ4)}
Figure~\ref{fig:abla_tos} compares ToS with CoS and ToT reasoning strategy under the same setting. ToS consistently performs best on all four metrics, including Iden. (46.77\%), Micro-F1 (25.84\%), Flip (44.26\%), and Trig (34.04\%). The gains are especially clear on the more structured and reasoning-intensive metrics, where ToS substantially outperforms both alternatives. These results show that the staged role-based design of ToS provides a more effective reasoning process for multimodal stance modeling and stance-flip analysis than sentiment-centered or generic tree-style prompting.

\begin{figure}[t!]  
	\centering    
	\includegraphics[width=\linewidth]{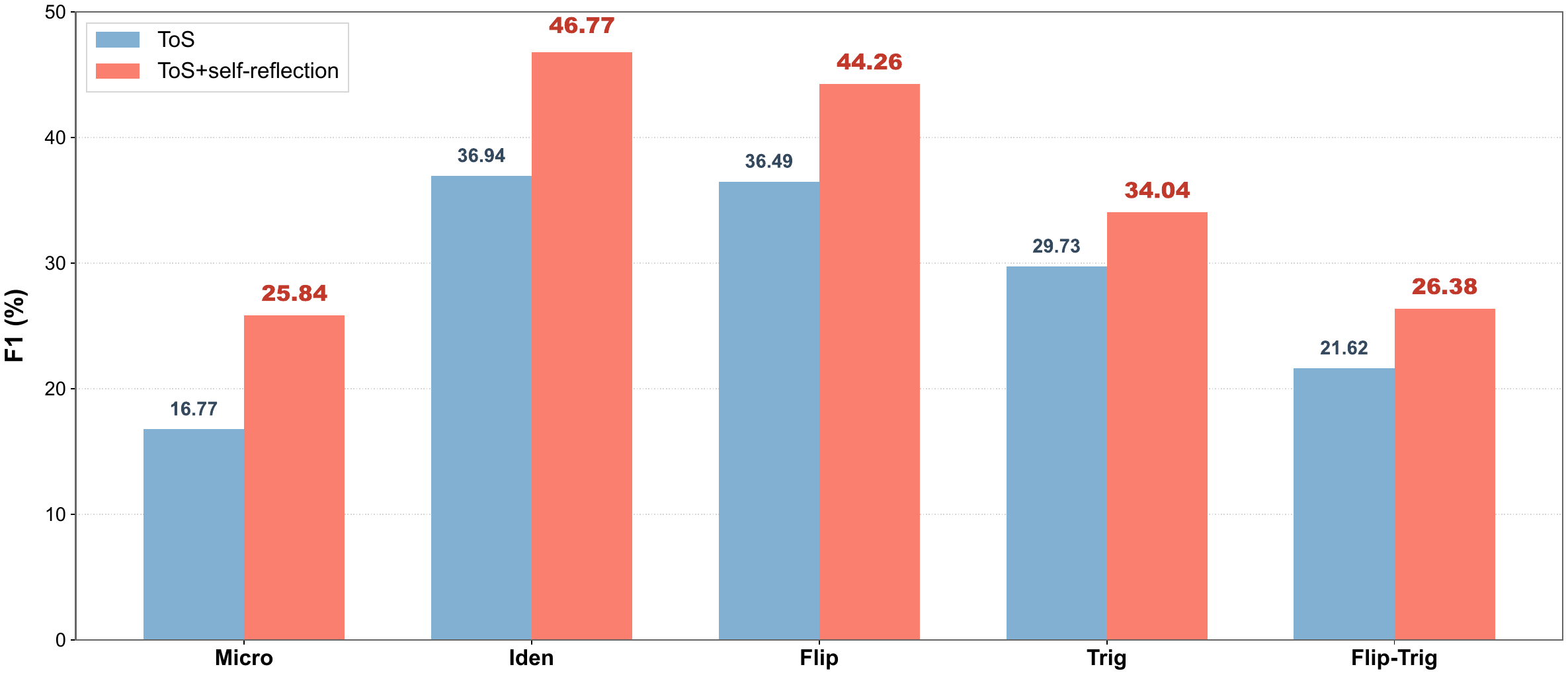} 
	\caption{Five-Dimensional Comparison: ToS (w/o Self-Reflection) vs ToS+Self-Reflection.}
	\label{fig:abla_self_reflection}
\end{figure}
\subsection{Impact of Multimodal Information (RQ5)}
Figure~\ref{fig:abla_multimodal} evaluates the contribution of each modality. The full ConStaFF model achieves the best performance on all metrics, reaching 46.77\% on Iden., 25.84\% on Micro-F1, and 26.38\% on Flip-Trig, confirming the benefit of joint multimodal modeling. Removing any modality leads to consistent degradation, but the impact is not uniform. Audio removal causes the largest drop on Iden. (42.76\%), suggesting that paralinguistic cues are important for fine-grained stance understanding, while removing video yields the lowest Flip-Trig score (22.58\%), indicating its particular value for identifying stance-reversal triggers. These results show that different modalities contribute complementary evidence, and that multimodal fusion is critical for both structural stance extraction and flip attribution.

\subsection{Impact of Verification Mechanism (RQ6)}
Figure~\ref{fig:abla_self_reflection} compares ToS with its full version augmented by self-reflection. Our proposed self-reflective verification mechanism consistently improves all five metrics. The largest improvements are observed in Micro and Iden., indicating that verification is particularly beneficial for structured prediction, while the consistent improvements on Flip, Trig, and Flip-Trig further show its value in refining evidence-grounded stance-flip reasoning. These results confirm that self-reflective verification mechanism improves both prediction reliability and explanation faithfulness.

\section{Conclusion}
This paper introduces multimodal conversational stance flipping forecasting, a novel task that advances stance detection by formalizing the cognitive evolution trajectory of human stance. It comprises two subtasks: (1) Multimodal Stance Sextuple Extraction, which provides a state-aware snapshot of each holder's conviction per turn through a structured record of holder, target, emotion, sentiment, stance, and rationale; and (2) Dynamic Stance Flip Attribution, which identifies the specific socio-cognitive trigger behind each belief reversal. We benchmark this novel setting with StanceFlip, a large-scale bilingual dataset built via a human-AI collaborative pipeline, covering five 
modalities, multi-turn multi-party contexts, high flip density, and expert-level annotation across diverse domains.  We further propose ConStaFF, a novel reasoning framework built on the thought-of-stance architecture and a self-reflective verification mechanism. By decomposing the complex reasoning process into specialized cognitive sub-tasks and mitigating causal hallucination through iterative rationale refinement, ConStaFF provides a strong baseline for future research on StanceFlip.



\bibliographystyle{ACM-Reference-Format}
\bibliography{refers}

\appendix
\section{More Details of Dataset}
\label{sec:appendix_data}

\subsection{Multi-stage Construction Pipeline}
\label{sec:appendix_pipeline}

\textbf{Seed Data Selection and Pre-processing}
To construct a benchmark that accurately reflects the stance evolution in online discourse, we implemented a rigorous data curation pipeline targeting high-quality textual seeds.

\textit{Source Corpus Integration and Rationale.} We aggregate raw dialogues from three sources: DailyDialog, MELD, and ZS-CSD. These datasets were chosen not only because they are conversational, but because they contain emotional conflict and argumentative density, both of which are useful for observable stance dynamics. Unlike generic chitchat corpora, they provide enough friction for opinion shifts to emerge naturally.

\textit{Heuristic Filtering for Stance Dynamics.} We find that stance evolution requires a specific temporal window to become visible. Dialogues shorter than 3 turns often lack enough context for a meaningful shift, whereas dialogues longer than 7 turns in existing corpora often suffer from topic drift or diluted context. We therefore apply a strict length filter of 3--7 turns. To keep the dataset rich in argumentative content, we also prioritize dialogues containing explicit discourse markers through keyword matching.

\textit{Speaker and Format Normalization.} Raw data from heterogeneous sources often use inconsistent schemas. We standardize all interlocutor labels into unified identifiers and normalize the conversation structure into a canonical JSON format. This pre-processing step reduces parsing errors in the later LLM-based simulation stage.

\textbf{Directed Multimodal Synthesis}
\label{sec:appendix_synthesis}
To transform textual dialogues into more realistic multimodal online chats, we design a \textit{Directed Multimodal Synthesis} pipeline. Rather than relying on random augmentation, the pipeline uses a multi-step reasoning strategy with GPT-4o. The detailed instructions that guide this analysis and generation process are shown in Table~\ref{box:prompt_analysis} and Table~\ref{box:prompt_generation}.

\begin{table}[h!]
    \centering
    \small
    \renewcommand{\arraystretch}{1.1}
    \begin{tcolorbox}[colback=gray!5!white,colframe=gray!75!black,title=\textbf{Prompt Box 1: Stance Analysis and Modality Planning Prompt}]
        ROLE: You are an expert analyst of online chat conversations. Your task is to identify key moments where users would naturally use multimodal elements.
        
        TASK A: Stance Significance Score. Rate from 0.0 to 1.0 how much this dialogue involves opinion, debate, or stance-taking.
        
        TASK B: Injection Point Recommendations. If the score is high ($>0.7$), identify potential \texttt{turn\_id}s using the following Priority Criteria:
        1) [Priority 1] Stance Shift Point: A clear change in a speaker's stance (e.g., realization, concession).
        2) [Priority 2] Stance Fortification: A moment of strong, emotional reinforcement of a stance.
        3) [Priority 3] Core Argumentation: A key fact or argument is presented.
        4) [Priority 4] Stance Exploration: A pivotal question that challenges another's stance.
        
        TASK C: Enhanced Online Chat Decision Tree.
        Q1: Visual Content? Does the intent involve showing proof? (Sharing photos, screenshots, memes?) $\rightarrow$ Choose IMAGE or VIDEO\_FRAME.
        Q2: Paralinguistic Audio? Does the intent involve sound? (Laughing "haha", crying, screaming, sound effects like "ding/wow") $\rightarrow$ Choose AUDIO.
        Q3: Emotional Reaction? Does the intent involve emotional expression? (Happy, sad, angry, surprised) $\rightarrow$ Choose EMOJI\_MEME or GIF.
        
        DIVERSITY BOOSTERS:
        1) Emotional Audio Priority: For strong emotions (laughing, yelling), strongly consider AUDIO first to capture intensity.
        2) Visual Variety: Don't default to memes; consider if IMAGE or VIDEO\_FRAME provides better evidence.
    \end{tcolorbox}
    \caption{The prompts for Step 1 (Significance Analysis) and Step 2 (Modality Selection), emphasizing diverse and logic-driven injection.}
    \label{box:prompt_analysis}
\end{table}

\textit{Step 1: Stance Significance Scoring and Pivot Identification.} As shown in Table~\ref{box:prompt_analysis}, we use a significance-driven injection strategy rather than a random one. The model first evaluates the full dialogue and assigns a Stance Significance Score (0.0--1.0). Only dialogues above a confidence threshold of 0.7 proceed to injection. To keep multimodal elements relevant to stance detection, the model selects insertion points according to a fixed priority order, giving preference to stance shifts and stance fortification over generic chatter. This helps ensure that each inserted item carries meaningful information about the speaker's conviction.

\textit{Step 2: Context-Aware Modality Selection (Decision Tree).} After identifying a target turn, the model follows the Enhanced Online Chat Decision Tree (Table~\ref{box:prompt_analysis}) to select the most appropriate modality. This logic distinguishes between visual evidence, such as screenshots used in argumentation, paralinguistic sounds such as laughter or sighs, and reactive imagery such as memes. We also include Diversity Boosters to prevent the model from defaulting to safe and repetitive choices like generic emojis, encouraging broader use of formats such as audio clips for intense emotion and video frames for dynamic reactions.

\begin{table}[h!]
    \centering
    \small
    \renewcommand{\arraystretch}{1.1}
    \begin{tcolorbox}[colback=gray!5!white,colframe=gray!75!black,title=\textbf{Prompt Box 2: Persona-based Query Generation Prompt (Irony-Aware)}]
        ROLE: You are a creative assistant. Transform a simple reason into a detailed, descriptive query.
        
        TASK: Generate a \texttt{retrieval\_query} that describes what the multimodal content should be.
        
        CRITICAL LOGIC (IRONY \& CONFLICT):
        1) Direct Match: If the speaker is sincere, the multimodal content should match the text emotion.
        2) Conflict Generation (Sarcasm/Irony): If the speaker is being sarcastic, passive-aggressive, or ironic, generate a description that contradicts the literal text.
        Example: Text = "Wow, you are a genius." (Sarcastic context) $\rightarrow$ Query = "A meme of a person facepalming or rolling their eyes in disbelief."
        
        REQUIREMENTS:
        1) Latent Intention: Describe the implied meaning and communicative goal, not just the literal text.
        2) Description Length: All descriptions MUST be 7-30 words long. Do not use short phrases.
        3) Specific Details: Include visual/audio details (e.g., lighting, expression intensity, specific sound textures).
        
        QUALITY CONTROL EXAMPLES:
        (Check) GOOD: "An animated sticker of a cartoon character turning red with anger, with steam coming out of its ears."
        (X) BAD: "Funny meme" (Too short/vague).
    \end{tcolorbox}
    \caption{The prompt for Step 3, featuring the specific mechanism for simulating sarcasm through text-image conflict.}
    \label{box:prompt_generation}
\end{table}

\textit{Step 3: Persona-based Query and Description Generation (Irony-Aware).} The final step bridges the gap between conversational context and multimodal retrieval. As detailed in Table~\ref{box:prompt_generation}, this step is important for simulating linguistic phenomena such as sarcasm. \textit{Text-Image Conflict Mechanism.} We explicitly instruct the model to analyze the speaker's latent intention. When sarcasm, irony, or passive aggression is present, the model generates visual descriptions that contradict the literal text, such as pairing ``Great idea'' with a ``Facepalm'' meme. This design forces downstream models to rely on multimodal reasoning instead of text alone. \textit{Rich Description Constraint.} To improve grounding quality, we impose a strict length constraint of 7--30 words and require sensory detail. This prevents vague queries like ``happy face'' and makes the generated captions more useful for retrieval from large-scale databases such as WebVid and AudioSet.

\textbf{Retrieval and Alignment}
\label{sec:appendix_retrieval}
After synthesizing descriptive queries, we run a retrieval process that maps those semantic descriptions to actual multimodal content. This stage addresses the challenge of aligning short conversational turns with large, weakly curated media repositories.

\textit{Multimodal Reservoirs and Pre-processing.} We build a large candidate pool by aggregating data from open-source collections. For static imagery, we use the full COCO dataset ($\sim$110k images) and a curated set of more than 15,000 expressive stickers. For temporal modalities, we use AudioSet and WebVid. A practical challenge is that the raw source files are much longer than online chat turns: videos and audio clips often last from 3 to more than 10 minutes, whereas the conversational events we model typically last only a few seconds.

\textit{Temporal Segmentation and Fine-grained Slicing.} To resolve this duration mismatch, we implement a \textit{Temporal Segmentation} pipeline. Instead of matching against full files, we pre-process raw audio and video into short segments, typically 3--8 seconds long, based on scene changes and sound event boundaries. This produces a large reservoir containing hundreds of thousands of candidate clips. As a result, retrieved media are not only semantically relevant but also temporally aligned with the pacing of a single dialogue turn.

\textit{Semantic Alignment and Diversity-Aware Sampling.} We use a dual-encoder framework (SentenceTransformer) to compute cosine similarity between the synthesized query and the captions of segmented clips. To keep the dataset from becoming repetitive, for example by repeatedly retrieving the same ``laughing'' clip, we adopt a Diversity-Aware Top-$k$ Sampling strategy with $k=10$. We retrieve the top-10 candidates above a dynamic similarity threshold and sample one at random. This preserves semantic fidelity while maintaining visual and auditory diversity.

\textbf{Automated Labeling}
\label{sec:appendix_labeling}
The final stage assigns fine-grained panoptic labels to every turn. We use GPT-4o as the annotator. Importantly, the model does not process text in isolation; it performs \textit{Cross-Modal Reasoning} to combine textual semantics with visual and auditory cues. The detailed labeling prompt is shown in Table~\ref{box:prompt_labeling}.

\begin{table}[h!]
    \centering
    \small
    \renewcommand{\arraystretch}{1.1}
    \begin{tcolorbox}[colback=gray!5!white,colframe=gray!75!black,title=\textbf{Prompt Box 3: Automated Panoptic Labeling and Multimodal Reasoning Prompt}]
        ROLE: You are a master of Discourse Analysis acting as a human conversation observer. Your task is to analyze the dialogue history and the provided multimodal captions to determine the speaker's true stance and emotion.
        
        INPUT DATA:
        1) Dialogue History: List of turns (Speaker, Text).
        2) Multimodal Description: The specific content of the image/audio attached to the current turn (e.g., "Image: A meme of a rolling eye").
        
        PRIMARY DIRECTIVE: CROSS-MODAL REASONING RULES.
        You must integrate the text and the multimodal description to form a final judgment.
        1) Conflict Resolution (Irony/Sarcasm): If the text is positive (e.g., "Great job") but the multimodal content present negative (e.g., "Image: Facepalm", "Audio: Boo sound"), you MUST prioritize the multimodal evidence. Label the sentiment as Negative and the intent as Sarcasm.
        2) Intensity Amplification: If the text is neutral but the multimodal content is high-arousal (e.g., "Audio: Screaming", "Image: Crying face"), update the Emotion label to reflect the high intensity (e.g., from "Neutral" to "Sadness" or "Anger").
        3) Visual Contextualization: Use the image content to resolve ambiguous pronouns (e.g., if text says "Look at this", and Image shows "A broken phone", the Topic/Target is "Phone Quality").
        
        ANALYTICAL FRAMEWORK: THE STANCE STATE MACHINE.
        1) Stance Persistence (Default): If the current turn is a filler, question, or ambiguous remark, the core stance remains UNCHANGED (inherits previous). shift\_occurred = "No".
        2) Stance Establishment: Moving from 'Unknown' to the FIRST declared stance is NOT a shift. It is Establishment. shift\_occurred = "No".
        3) True Stance Shift (Reversal): The ONLY case for "Yes". Requires moving from one ESTABLISHED stance (e.g., Support) to a DIFFERENT ESTABLISHED stance (e.g., Oppose).
        
        OUTPUT SCHEMA (JSON):
        - keywords: [Array of 3-5 strings capturing essence]
        - emotion: [Selected from predefined Emotion Labels]
        - sentiment: [Positive, Negative, Neutral]
        - stance\_after\_shift: [Support, Oppose, Neutral]
        - stance\_reason\_detail: "Explanation citing specific Text AND Multimodal evidence."
        - shift\_occurred: "Yes" or "No"
        - shift\_reason\_category: [Introduction of New Info, Logical Argument, Emotional Appeal, Social Pressure]
    \end{tcolorbox}
    \caption{The detailed system prompt for the labeling phase, explicitly enforcing multimodal integration and stance state logic.}
    \label{box:prompt_labeling}
\end{table}

\textit{Multimodal Synergy in Annotation.} As detailed in the ``Cross-Modal Reasoning Rules'' of Table~\ref{box:prompt_labeling}, the prompt explicitly handles interactions across modalities. We instruct the model to prioritize multimodal evidence when it conflicts with text, as in sarcasm, and to use it to refine emotional intensity, for example by upgrading ``displeasure'' to ``anger'' based on audio cues. This helps ensure that the labels reflect the \textit{holistic} communicative intent rather than the textual surface form alone.

\textit{Formalized Stance Evolution Model.} To ensure rigorous stance tracking, we embed a ``Stance State Machine'' inside the prompt. This makes a strict distinction between \textit{Stance Establishment} (initial opinion formation) and \textit{True Stance Shift} (polarity reversal). By enforcing these definitions, we filter out noise such as temporary hesitation or off-topic diversion, so that positive \texttt{shift\_occurred} labels correspond to genuine cognitive change.

\textbf{Human Verification Protocols}
\label{sec:appendix_qa}
The entire dataset was assigned to a panel of 17 expert annotators (postgraduate students in NLP) for comprehensive turn-by-turn verification. Rather than simply relabeling, the annotators performed targeted refinement and filtering according to the following criteria:
\begin{itemize}[leftmargin=1.5em]
    \item \textbf{Target and Stance Alignment:} Annotators inspected the logical consistency between the textual utterances and the assigned stance labels. If a stance flip point identified by the pipeline was found to be ambiguous or lacked a clear causal trigger after expert deliberation, the sample was flagged and discarded.
    \item \textbf{Contextual Media Verification:} Experts reviewed the multimodal content associated with each turn. If a media item (image, audio, or video) was identified as semantically misaligned with the dialogue context or the speaker's intent, it was manually reset to \texttt{N/A} to maintain data fidelity.
    \item \textbf{Heuristic Correction of Stance Inheritance:} Following the \textit{Stance Persistence Rule}, annotators corrected instances where the pipeline failed to propagate the previous definitive stance ($st_{prev}$) to filler or off-target turns, ensuring the continuous trajectory of speaker convictions.
    \item \textbf{Three-Strike Filtering Protocol:} During the manual audit, a "Three-Strike" policy was enforced for terminal error handling. A dialogue was permanently removed from the benchmark if: (1) the central proposition $G$ remained elusive despite manual review; (2) consensus on the stance transition could not be reached among experts; or (3) the rationale was found to be logically irrecoverable.
\end{itemize}

After the audit and refinement phase, we conducted a validation study to measure the objectivity of the final dataset:
\begin{itemize}[leftmargin=1.5em]
    \item \textbf{Sampling and Protocol:} A random subset of 100 dialogues ($\sim$600 turns) was extracted from the refined pool. Two senior annotators independently re-labeled this subset in a double-blind manner, without access to previous pipeline outputs or each other's decisions.
    \item \textbf{Statistical Reliability:} Inter-annotator agreement was measured using Cohen's Kappa ($\kappa$). The analysis yielded $\kappa = 0.84$ for stance identification and $\kappa = 0.87$ for multimodal relevance. These high scores confirm that the \textit{StanceFlip} annotation guidelines are highly objective and the final labels are consistent.
\end{itemize}

\subsection{ Detailed Summary of Dataset Insights}
\label{sec:data_insights}

$\blacklozenge$ \textbf{Panoptic Sextuple: The Structural Decoupling of Affect and Argument.}
We argue that in multi-party discourse, the binary ``Support/Oppose'' paradigm is insufficient because \textit{Affect} (Sentiment) and \textit{Argumentation} (Stance) operate on different cognitive dimensions. Treating them as the same signal is one of the main sources of noise in existing benchmarks.
(1) \textit{Target-Specific Disambiguation:} In complex debates, a holder ($h$) often directs negative sentiment towards an opponent's \textit{tone} or \textit{personality} (ad hominem attacks) while explicitly supporting their \textit{logical premise} ($t$). A monolithic label fails to distinguish "interpersonal hostility" from "topical disagreement." Our Panoptic Sextuple $(h, t, e, s, st, r)$ resolves this ambiguity by strictly binding the holder's affective state ($e, s$) and stance ($st$) to the same specific target ($t$). This structural decoupling ensures that the model captures the precise cognitive object of the user's expression, preventing the misclassification of "hostile agreement" as opposition.
(2) \textit{Rationale as Structural Evidence:} We define the Rationale ($r$) not merely as a supplementary text span, but as the grounding evidence that validates the stance. It forces the model to ground its prediction in specific discursive logic rather than spurious lexical correlations (e.g., blindly classifying all turns containing "but" as opposition). Thus, the sextuple serves as the minimal sufficient statistic required to represent a static cognitive state without ambiguity.

$\blacklozenge$ \textbf{Lifecycle Dynamics: Modeling the Inertia of Belief.}
Stance is not a discrete event triggered at every utterance, but a continuous cognitive state shaped by \textit{Cognitive Inertia}. Existing methods often treat each turn as an independent classification task and ignore the strong temporal dependencies in dialogue. We introduce a formalized Stance State Machine to capture that continuity:
(1) \textit{Establishment} ($S_{\emptyset} \to S_{A}$): The initial formation of an opinion from an unknown state.
(2) \textit{Persistence} ($S_{A} \to S_{A}$): Subsequent turns where the holder reiterates, clarifies, or defends the existing view. Crucially, explicitly modeling persistence allows the system to filter out "false flips" caused by mere conversational elaboration or rhetorical hesitation.
(3) \textit{True Flipping} ($S_{A} \to S_{B}$): A low-frequency, high-energy cognitive rupture where the holder's internal state is rewritten. This dynamic perspective transforms the task from simple classification to complex state tracking, requiring models to distinguish between the maintenance of an existing belief and the genuine adoption of a new one.

$\blacklozenge$ \textbf{Causal Heptatuple: From Descriptive State to Explanatory Mechanism.}
The occurrence of a \textit{True Flip} marks a qualitative change in the data structure, extending the static Sextuple into a dynamic Heptatuple by adding a seventh element: the \textit{Trigger} ($\tau$).
While the sextuple serves as a \textit{Descriptive Snapshot} of what the current cognitive state is, it lacks the explanatory power to define the transition mechanism. The Heptatuple functions as a \textit{Causal Event Record}, where the Trigger ($\tau$) represents the specific \textit{External Force}—such as Logical Correction, Emotional Resonance, or Social Pressure—that possessed sufficient energy to overcome the holder's Stance Inertia. This transition effectively models the persuasion process as an "Input-Output" system, where $\tau$ is the stimulus and the stance flip is the mandatory response. This structure allows StanceFlip to benchmark not just the detection of opinion shifts, but the attribution of their socio-cognitive causes.

$\blacklozenge$ \textbf{Taxonomy: Deep Socio-Cognitive Triggers.}
We categorize persuasion mechanisms ($\tau$) into four cognitive dimensions to capture the ``why'' behind each flip:
(1) \textit{Logical Argumentation} (Epistemic Change): The flip is driven by factual correction, reasoning, or the exposure of logical fallacies. This represents a change in knowledge or rationality.
(2) \textit{Emotional Resonance} (Affective Alignment): The flip is triggered by empathy, anger, or tonal alignment (e.g., "I feel your pain"). This represents a change driven by emotional contagion rather than facts.
(3) \textit{New Information} (Contextual Update): The flip occurs because new external evidence (news, data) is introduced that changes the premise of the debate.
(4) \textit{Social Interaction} (Group Dynamics): The flip is a result of peer pressure, the desire for consensus, or deference to authority. This represents a socially motivated compliance.

$\blacklozenge$ \textbf{Authentic Multimodal Context.} To reduce the ``synthetic bias'' common in machine-generated corpora, \textit{StanceFlip} grounds its textual base in high-quality human conversation datasets. This helps preserve real conversational phenomena such as colloquial language and complex turn-taking. Within this setting, we implement a three-stage causal synthesis pipeline so that non-textual media carry meaningful \textit{pragmatic weight}. The process begins with \textit{strategic trigger localization}, which identifies the turn where a stance flip occurs as the logical anchor. It then uses \textit{contextual modality selection} to choose the most effective medium, such as a cynical audio clip or a counter-evidential image. Finally, \textit{semantic alignment via captioning} produces high-fidelity rationales that connect visual or auditory cues with the dialogue flow, ensuring that multimodal information act as genuine drivers of stance evolution.

$\blacklozenge$ \textbf{Multi-domain Diversity.} To evaluate model generalizability rigorously, \textit{StanceFlip} spans ten major domains, from high-stakes areas such as \textit{Politics} to everyday settings such as \textit{Relationship}. This breadth matters for benchmarking \textit{Stance Inertia}, because convictions in social or ethical domains often resist change more strongly than consumer preferences. By covering more than 100 sub-domains with substantial argumentative friction, the dataset makes it harder for models to rely on domain-specific shortcuts. The result is a broader test of adaptability across discursive structures and themes.

\begin{figure}[t] 
    \centering
    \includegraphics[width=\linewidth]{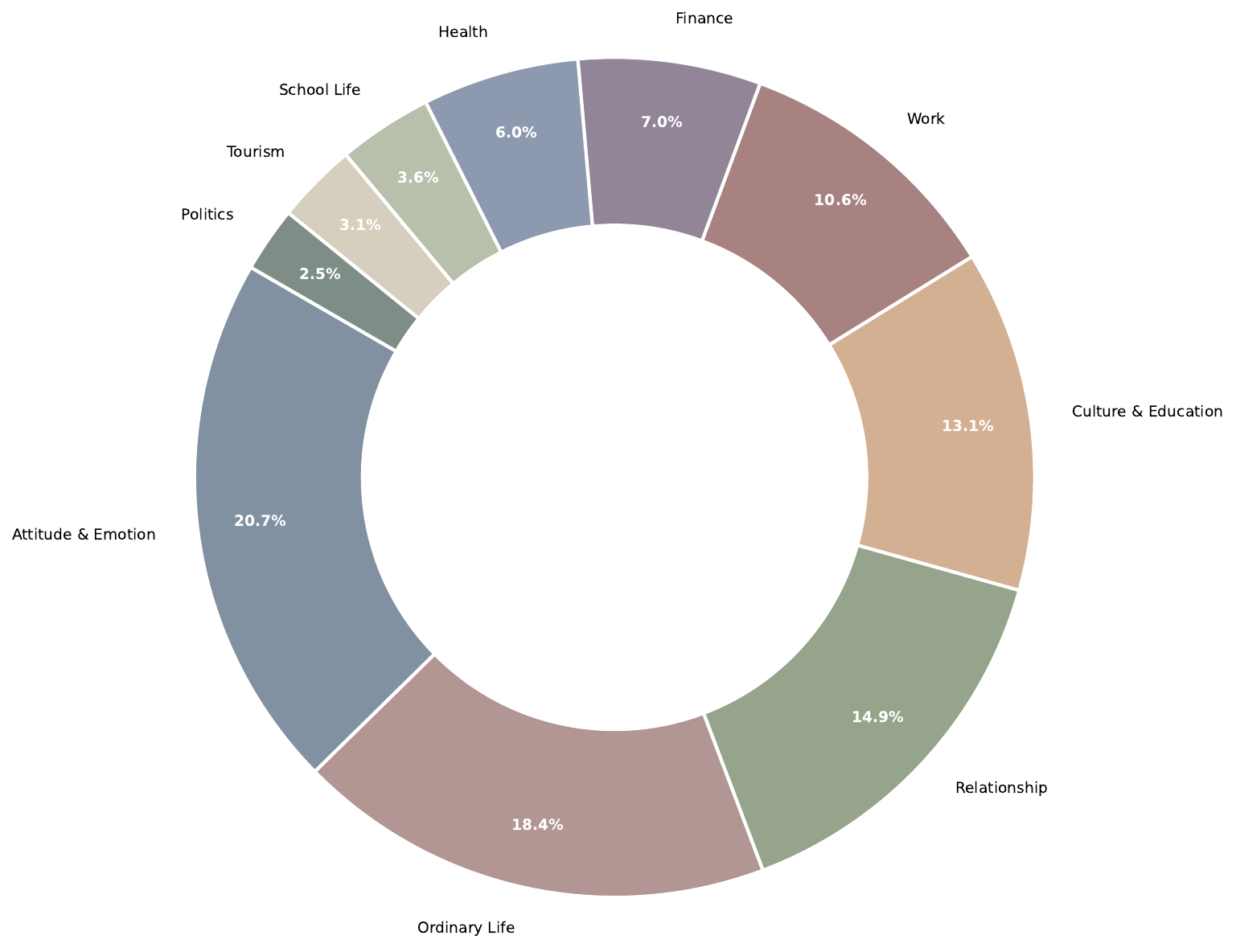} 
    \Description{A donut chart illustrating the distribution of principal domains in the StanceFlip dataset, highlighting thematic breadth.}
    \caption{Distribution of principal domains in the dataset.}
    \label{fig:topic_dist}
\end{figure}

$\blacklozenge$ \textbf{Human-AI Collaborative Annotation Pipeline.} Building a large-scale dataset with complex reasoning labels requires a balance between automated scalability and expert review. Our \textit{Human-in-the-loop} workflow uses LLMs for labor-intensive preprocessing tasks such as multimodal retrieval and draft rationale generation. To maintain annotation fidelity, every instance then passes through a strict audit protocol carried out by 17 NLP experts. The experts focus on validating the \textit{logical entailment} between the assigned stance and its rationale, ensuring that each transition is supported by a verifiable cognitive chain. We further apply a ``Three-Strike'' filtering rule to discard dialogues with unclear goals or irrecoverable logic, resulting in a high Cohen’s Kappa ($\kappa = 0.84$).

\section{More Details of Methods}
\label{appendix:methods}

\subsection{Prompting Design for ToS Reasoning framework}
To ensure reproducibility and facilitate further research, we provide the detailed prompt templates used in our ToS reasoning framework. The framework utilizes a step-by-step persona-based generation process.

\subsubsection{Step 1: The Cartographer (Target Proposition Identification)}
\textbf{Goal:} Perform a global scan of the dialogue to formulate a consistent debate proposition.

\textbf{Prompt Template:}
\begin{itemize}
    \item \textbf{Instruction:} Act as a \textbf{Cartographer}. Your primary goal is to identify the central 'Target' of this entire conversation. To do this:
    \begin{itemize}
        \item \textit{Global Scan:} Scan the entire dialogue to identify the main topic of contention, the central proposal, or the core issue being debated.
        \item \textit{Formulate Proposition:} Convert this identified core issue into a clear, concise debate proposition sentence, using constructions like 'The debate concerning whether...' or 'The discussion about whether...'. The 'Target' \textbf{MUST} be phrased as a precise debate proposition, not a simple noun (e.g., instead of 'coffee', frame it as 'The debate concerning whether coffee should be tried as a substitute for cigarettes'). This proposition will serve as the consistent reference point for all stances in this dialogue.
    \end{itemize}
\end{itemize}

\textbf{Input Data:}
\begin{itemize}
    \item Full Dialogue: \{full\_dialogue\_text\_with\_captions\}
\end{itemize}

\textbf{Output:}
\begin{itemize}
    \item (Target: [Debate Proposition Sentence])
\end{itemize}

\subsubsection{Step 2: The Psychologist (Multimodal Conflict Resolution)}
\textbf{Goal:} Analyze the sentiment and emotion of a specific turn, resolving conflicts between text and visual cues.

\textbf{Prompt Template:}
\begin{itemize}
    \item \textbf{Instruction:} Act as a \textbf{Psychologist} and \textbf{Emotion Analyst}. Analyze \{holder\_name\}'s current utterance, including its textual content and the provided multimodal cue descriptions, to determine its core Emotion and Sentiment.
    \begin{itemize}
        \item \textit{Textual Layer:} Begin by analyzing what the utterance explicitly states and identifying initial emotion and sentiment solely based on the text.
        \item \textit{Multimodal Integration:} Carefully examine the 'multimodal cue descriptions' to see if these non-verbal cues (e.g., described facial expressions, tone) reinforce, contradict, or enrich the textual meaning.
        \item \textit{Conflict Resolution:} If there is a conflict between textual and multimodal cues (e.g., if the text says 'That's great' but the multimodal cue describes 'a sarcastic smile'), prioritize interpretations that resolve ambiguity (like detecting sarcasm) to reveal \{holder\_name\}'s true feelings.
        \item \textit{Final Selection:} Choose the Emotion and Sentiment from the allowed lists that best reflect \{holder\_name\}'s attitude towards \{gt\_target\}.
    \end{itemize}
\end{itemize}

\textbf{Input Data:}
\begin{itemize}
    \item Dialogue History: \{history\_up\_to\_current\_turn\}
    \item Current Speaker: \{holder\_name\}
    \item Utterance \& Cues: \{current\_utterance\_with\_caption\}
    \item Target: \{gt\_target\}
    \item Constraints:
    \begin{itemize}
        \item Emotions=\{EMOTION\_LABELS\}
        \item Sentiments=\{SENTIMENT\_LABELS\}
    \end{itemize}
\end{itemize}

\textbf{Output:}
\begin{itemize}
    \item (Holder: [Name], Target: [Proposition], Sentiment: [Class], Emotion: [Class])
\end{itemize}

\subsubsection{Step 3: The Discourse Analyst (Stance Determination)}
\textbf{Goal:} Determine the final stance by comparing current sentiment with the historical stance trajectory.

\textbf{Prompt Template:}
\begin{itemize}
    \item \textbf{Instruction:} Act as a \textbf{Discourse Analyst}. Based on the complete dialogue history (especially all prior utterances of \{holder\_name\} and their multimodal cues), the current sentiment/emotion, and \{holder\_name\}'s 'Last Stance', determine \{holder\_name\}'s final Stance towards \{gt\_target\}.
    \begin{itemize}
        \item \textit{Historical Review:} Carefully review all prior utterances of \{holder\_name\} regarding \{gt\_target\}, focusing on the evolution of their expressed content and emotions.
        \item \textit{Temporal Comparison:} Evaluate the current utterance's sentiment (\{gt\_sentiment\}) and emotion (\{gt\_emotion\}) by comparing it with \{holder\_name\}'s previous utterances and 'Last Stance' (\{previous\_stance\_for\_holder\}). Determine if it is consistent, slightly divergent, or a clear shift.
        \item \textit{Final Decision:} Choose the Stance that best reflects \{holder\_name\}'s current stance. Select 'Support' or 'Oppose' only when the opinion is clear; otherwise, choose 'Neutral' or 'Unknown'.
    \end{itemize}
\end{itemize}

\textbf{Input Data:}
\begin{itemize}
    \item Dialogue History: \{history\}
    \item Current Sentiment/Emotion: (Sentiment: \{gt\_sentiment\}, Emotion: \{gt\_emotion\})
    \item Last Stance: \{previous\_stance\_for\_holder\}
\end{itemize}

\textbf{Output:}
\begin{itemize}
    \item (Holder: [Name], Target: [Proposition], Stance: [Class])
\end{itemize}

\subsubsection{Step 4: The Synthesizer \& Critic (Rationale Generation and Reflexion)}
\textbf{Goal:} Generate a logic-based rationale and perform self-correction.

\textbf{Prompt Template:}
\begin{itemize}
    \item \textbf{Instruction:} Act as a \textbf{Synthesizer} and \textbf{Critic}.
    \begin{itemize}
        \item \textit{Draft Rationale (Synthesizer):} Based on the 'Intermediate Analysis Results' and 'Dialogue History', construct a concise 'Rationale' (1-2 sentences) that logically explains WHY \{holder\_name\} holds this \{gt\_stance\}. Reference specific textual evidence. If multimodal cues were crucial, explicitly mention their description (e.g., 'despite positive words, their sarcastic facial expression indicated opposition').
        \item \textit{Self-Reflexion (Critic):} Critically evaluate the draft: Does it logically, clearly, and powerfully support the stance? Is it specific enough? If the rationale feels weak or unconvincing, revise it to provide a stronger, more precise explanation.
    \end{itemize}
\end{itemize}

\textbf{Input Data:}
\begin{itemize}
    \item Stance Info: (Speaker: \{holder\_name\}, Target: \{gt\_target\}, Stance: \{gt\_stance\})
    \item Constraints: Rationale \textbf{MUST} be 1-2 sentences.
\end{itemize}

\textbf{Output:}
\begin{itemize}
    \item (Holder: [Name], ..., Stance: [Class], Rationale: [Analytical Summary])
\end{itemize}

\subsection{Self-Reflection Instruction Tuning Implementation}
To train the metacognitive capability required by the self-reflective mechanism in Section 4.3, we construct a specialized instruction-tuning dataset. Rather than relying only on inference-time prompting, we explicitly train the model to criticize and correct errors through a ``Corrupt-and-Correct'' data augmentation strategy.

The construction process generates three types of training tasks from the ground-truth dataset:

\subsubsection{Data Construction Logic}
We implemented a \texttt{SelfRefineDataset} class that processes the raw StanceFlip dialogues. For every ground truth stance tuple in the dataset, we apply the following logic:

\textbf{Task 1: Initial Generation (Standard Supervised Fine-Tuning)}
\begin{itemize}
    \item \textbf{Input:} The full dialogue history.
    \item \textbf{Instruction:} "Extract all stance information from the dialogue."
    \item \textbf{Target:} The correct, ground-truth stance sextuples.
\end{itemize}

\textbf{Task 2: Feedback Generation (Criticism)}
\begin{itemize}
    \item \textit{Objective:} To teach the model to identify errors, we artificially "corrupt" the ground truth tuples.
    \item \textbf{Corruption Strategy:} We verify specific elements of the tuple based on a random selection:
    \begin{itemize}
        \item Holder Corruption: Swapping the true speaker with another participant in the dialogue.
        \item Stance Corruption: Inverting the stance label (e.g., changing "Support" to "Oppose") or modifying the "Flipped Stance" label in transition samples.
        \item Target Corruption: Replacing the specific proposition with a vague placeholder (e.g., "something").
    \end{itemize}
    \item \textbf{Instruction:} "Provide specific feedback on the extracted stance tuple based on the dialogue."
    \item \textbf{Input:} The dialogue + The Corrupted Tuple.
    \item \textbf{Target:} A natural language explanation of the error (e.g., "The Stance is incorrect. The speaker's attitude is 'Support', not 'Oppose'.").
\end{itemize}

\textbf{Task 3: Output Refinement (Correction)}
\begin{itemize}
    \item \textbf{Instruction:} "Fix the extracted stance tuple based on the feedback."
    \item \textbf{Input:} The dialogue + The Corrupted Tuple + The Feedback from Task 2.
    \item \textbf{Target:} The original, correct Ground Truth tuple.
\end{itemize}

\subsubsection{Training Specifications}
\begin{itemize}
    \item \textbf{Template:} All prompts are wrapped in a standard chat template (e.g., Vicuna format: USER: ... ASSISTANT: ...) to align with the backbone LLM's instruction format.
    \item \textbf{Negative Sampling:} To maintain data balance and prevent the model from overfitting to correction tasks, we apply a down-sampling rate (probability > 0.4) to the self-reflection samples.
    \item \textbf{Sequence Construction:} The input sequences are tokenized with specific handling for EOS (End of Sentence) tokens to ensure the model learns to terminate generation correctly. We utilize masking (setting labels to -100) on the prompt instructions so the model is only trained on the Target outputs (the extraction, the feedback, and the correction).
\end{itemize}

\section{Extensions of Settings and Implementations}
\label{sec:appendix_settings}

To support reproducibility, we provide detailed specifications of the computational infrastructure, model architecture, and hyperparameter settings used in the three-stage curriculum learning pipeline.

\subsection{Computational Infrastructure}
All experiments were conducted on a high-performance computing cluster with \textbf{10 $\times$ NVIDIA RTX A6000 GPUs} (48GB VRAM per GPU). The software environment used \textbf{PyTorch 2.3.1}, \textbf{Hugging Face Transformers 4.57.1}, \textbf{PEFT 0.18.0}, and \textbf{Accelerate 1.12.0}. To improve memory efficiency during distributed training, we used \textbf{DeepSpeed (v0.18.2)} with ZeRO-2 optimization and gradient checkpointing. We also used \textbf{BitsAndBytes (v0.48.2)} for 4-bit and 8-bit quantization, and NCCL for cross-GPU communication.

\subsection{Model Architecture Configurations}

\paragraph{Backbone and Visual Encoder.} 
We use \textbf{Vicuna-7B-v1.5}~\cite{Vicuna} as the Large Language Model (LLM) backbone because of its strong instruction-following ability. For visual perception, we use the frozen \textbf{ImageBind-Huge}~\cite{ImageBind} encoder. ImageBind produces a fixed 1024-dimensional embedding for inputs across modalities, including image, video, and audio.

\paragraph{Modality Projector.} 
To map visual embeddings into the LLM input space (4096 dimensions for Llama-2/Vicuna-7B), we design a learnable Multi-Layer Perceptron (MLP) projector. As used in Stage 1, the projector is defined as:
\begin{equation}
\text{Projector}(x) = W_2(\text{GELU}(\text{LayerNorm}(W_1 x)))
\end{equation}
where $W_1 \in \mathbb{R}^{1024 \times 4096}$ and $W_2 \in \mathbb{R}^{4096 \times 4096}$. During training, all visual inputs are normalized prior to projection.

\paragraph{Parameter-Efficient Fine-Tuning (PEFT).} 
We use Low-Rank Adaptation (LoRA) in Stages 2 and 3. To balance performance and memory use on A6000 GPUs, we apply QLoRA-style quantization strategies:
\begin{itemize}
    \item \textbf{Stage 2:} 8-bit quantization (\texttt{load\_in\_8bit=True}).
    \item \textbf{Stage 3:} 4-bit Normal Float (NF4) quantization with double quantization enabled (\texttt{bnb\_4bit\_quant\_type="nf4"}).
\end{itemize}
The LoRA adapters were attached to all linear layers: \texttt{q\_proj}, \texttt{k\_proj}, \texttt{v\_proj}, \texttt{o\_proj}, \texttt{gate\_proj}, \texttt{up\_proj}, and \texttt{down\_proj}.

\subsection{Training Protocols by Stage}

We adopt a progressive three-stage training pipeline. The detailed hyperparameters for each stage are summarized in Table~\ref{tab:hyperparams}.

\paragraph{Stage 1: Multimodal Alignment (Perception).}
In this stage, we freeze the LLM backbone and the ImageBind encoder, and train \textit{only} the Projector. The objective is to align visual features with the text embedding space using image-caption pairs from our dataset. We optimize this stage with AdamW and a learning rate of 2e-5.

\paragraph{Stage 2: ToS Reasoning Tuning (Cognition).}
We freeze the pre-trained Projector and the ImageBind encoder, and enable gradients \textit{only} for the LoRA adapters. The model is then trained on the Panoptic Sextuple Extraction task. We use a larger effective batch size through gradient accumulation ($steps=32$) to stabilize the learning of longer reasoning chains.

\paragraph{Stage 3: Self-Reflection Tuning (Metacognition).}
In the final stage, we continue fine-tuning the LoRA adapters initialized from Stage 2 while keeping the Projector frozen. To reduce the instability often seen in 4-bit training, we use the \texttt{paged\_adamw\_32bit} optimizer and gradient clipping (max norm 0.3). This stage focuses on the ``Corrupt-and-Correct'' objective so that the model becomes better at revising hallucinated reasoning.

\begin{table}[h]
\centering
\caption{Hyperparameter settings for the three training stages of ConStaFF.}
\label{tab:hyperparams}
\resizebox{0.95\columnwidth}{!}{%
\begin{tabular}{l|ccc}
\toprule
\textbf{Hyperparameter} & \textbf{Stage 1} & \textbf{Stage 2} & \textbf{Stage 3} \\
\midrule
Trainable Parameters & Projector & LoRA Adapters & LoRA Adapters \\
LLM Quantization & FP16 & Int8 & NF4 (4-bit) \\
LoRA Rank ($r$) & N/A & 16 & 16 \\
LoRA Alpha ($\alpha$) & N/A & 32 & 32 \\
LoRA Dropout & N/A & 0.05 & 0.05 \\
Optimizer & AdamW & AdamW & Paged AdamW 32bit \\
Learning Rate & 2e-5 & 1e-4 & 5e-5 \\
LR Scheduler & Cosine & Linear & Linear \\
Warmup Ratio & 0.0 & 0.03 & 0.03 \\
Per-Device Batch Size & 4 & 1 & 4 \\
Grad. Accumulation & 1 & 32 & 16 \\
Max Grad Norm & 1.0 & 1.0 & 0.3 \\
Epochs & 3 & 2 & 2 \\
\bottomrule
\end{tabular}%
}
\end{table}

\subsection{Training Objectives}
Across all three stages, the training objective is to maximize the log-likelihood of target tokens conditioned on the input context and multimodal features. 

Formally, let $X = \{x_1, x_2, \dots, x_N\}$ be the input sequence (including the visual embeddings $v$ mapped by the projector) and $Y = \{y_1, y_2, \dots, y_M\}$ be the target response sequence. The standard autoregressive language modeling loss $\mathcal{L}_{CLM}$ is defined as:

\begin{equation}
    \mathcal{L}_{CLM}(\theta) = - \sum_{t=1}^{M} \log P_\theta(y_t \mid X, y_{<t})
\end{equation}

where $\theta$ represents the trainable parameters (Projector in Stage 1, LoRA adapters in Stages 2/3).

\paragraph{Stage-Specific Objectives.}
\begin{itemize}
    \item \textbf{Stage 1 (Multimodal Alignment):} $X$ consists of the frozen image features and a prefix prompt, while $Y$ is the ground-truth caption. The loss ensures the projector aligns visual concepts with textual semantics.
    
    \item \textbf{Stage 2 (ToS Reasoning):} $X$ is the dialogue history with interleaved visual cues, and $Y$ is the Chain-of-Thought reasoning path (Cartographer $\to$ Psychologist $\to$ Analyst). We apply a masking strategy where the loss is calculated \textit{only} on the reasoning steps and final stance labels, ignoring the instruction prompt tokens ($label = -100$).
    
    \item \textbf{Stage 3 (Self-Reflection):} This stage optimizes a multi-task objective. The model learns to generate both the critique ($Y_{feedback}$) and the correction ($Y_{refined}$). The total loss is the summation over the self-correction samples:
    \begin{equation}
        \mathcal{L}_{Stage3} = \mathcal{L}_{CLM}(Y_{feedback} \mid X_{draft}) + \mathcal{L}_{CLM}(Y_{refined} \mid X_{draft}, Y_{feedback})
    \end{equation}
\end{itemize}

\section{Evaluation Specifications}
\label{sec:appendix_eval}

Because Large Language Models are generative, exact string matching is often too strict for evaluating complex reasoning tasks. We therefore use a multi-layer evaluation protocol that combines rule-based normalization with semantic evaluation through an LLM-as-a-Judge.

\subsection{Preprocessing and Normalization}
Prior to evaluation, all predicted ($y_{pred}$) and ground-truth ($y_{gold}$) outputs undergo a standard normalization process $\mathcal{N}(\cdot)$, which includes:
\begin{itemize}
    \item \textbf{Case Folding \& Whitespace:} Converting all text to lowercase and stripping leading/trailing whitespace.
    \item \textbf{Holder Alias Resolution:} Mapping variations of speaker identifiers to a canonical form (e.g., ``\textit{Speaker 1}'', ``\textit{s1}'', ``\textit{user 1}'' $\rightarrow$ ``\textit{speaker 1}'').
\end{itemize}

\subsection{Rule-based Categorical Matching}
For fields with a closed label set, we employ heuristic keyword matching to handle lexical variations in generation.

\paragraph{Stance Mapping.} 
We verify if the predicted stance falls into the correct polarity bucket:
\begin{itemize}
    \item \textbf{Support:} Contains \{``support'', ``agree'', ``pro'', ``positive''\}.
    \item \textbf{Oppose:} Contains \{``oppose'', ``disagree'', ``con'', ``negative'', ``refus''\}.
    \item \textbf{Neutral:} Contains \{``neutral''\}.
\end{itemize}

\paragraph{Trigger Categorization.} 
Since the model generates free-text triggers, we map them to the four taxonomy categories defined in Section 3.1 using keyword heuristics:
\begin{itemize}
    \item \textbf{Factual \& Logical:} \{``fact'', ``logic'', ``evidence'', ``data'', ``stat''\}.
    \item \textbf{Emotional \& Value:} \{``emotion'', ``value'', ``moral'', ``empathy'', ``fear''\}.
    \item \textbf{Personal Experience:} \{``experi'', ``story'', ``anecdote'', ``life''\}.
    \item \textbf{Social Influence:} \{``social'', ``pressure'', ``group'', ``norm'', ``peer''\}.
\end{itemize}
A prediction is a True Positive if its mapped category matches the ground truth.

\subsection{Model-based Semantic Evaluation}
For open-ended fields (\textbf{Target} and \textbf{Rationale}), rigid string matching yields high false negatives. We adopt an \textbf{LLM-as-a-Judge} approach using \texttt{GPT-4o-mini}.

We construct a dynamic prompt $\mathcal{P}_{eval}$ incorporating the full dialogue context $\mathcal{C}$. The judge is instructed to output ``YES'' only if semantic equivalence is met. The prompt template used is:

\begin{quote}
\small
\textbf{System:} You are a helpful assistant for evaluating text similarity. Respond ONLY with 'YES' or 'NO'.\\
\textbf{User:} Dialogue Context: [$\mathcal{C}$] \\
Predicted Target: [$y_{pred}$] \\
Gold Target: [$y_{gold}$] \\
Is the Predicted Target semantically equivalent to, contained within, or does it refer to the core entity/proposition of the Gold Target?
\end{quote}

To ensure reproducibility and efficiency, we implement a persistent caching mechanism (saved as \texttt{api\_match\_cache.json}). If the API call fails or times out, the system falls back to a lenient string containment check (i.e., match if $y_{pred} \subset y_{gold}$ or $y_{gold} \subset y_{pred}$).

\subsection{Compound Metric Definitions}
We adopt the standard Precision ($P$), Recall ($R$), and F1-score ($F1$) as our primary evaluation metrics. Based on the cumulative count of True Positives ($TP$), False Positives ($FP$), and False Negatives ($FN$) across the entire test set, the scores are calculated as follows:

\begin{equation}
    P = \frac{TP}{TP + FP}, \quad R = \frac{TP}{TP + FN}, \quad F1 = \frac{2 \cdot P \cdot R}{P + R}
\end{equation}

To capture different granularities of model performance, we define specific tuple structures for matching criteria:

\begin{itemize}
    \item \textbf{Panoptic Micro F1:} Requires the correct extraction of the complete sextuple: \textit{(Holder, Target, Emotion, Sentiment, Stance, Rationale)}. This is the strictest metric.
    \item \textbf{Identification (Iden) F1:} Focuses on the "who" and "why", requiring the tuple: \textit{(Holder, Target, Rationale)}.
    \item \textbf{Flip-Trig F1:} The core metric for our StanceFlip task. It evaluates the dynamic transition logic, requiring the tuple:
    \begin{equation}
        (Stance_{initial}, Stance_{flipped}, Trigger_{causal})
    \end{equation}
    This metric verifies that the model has correctly identified \textit{both} the stance reversal event and its underlying cause.
\end{itemize}

\end{document}